\def\year{2021}\relax
\documentclass[letterpaper]{article} 
\usepackage{aaai21}  
\usepackage{times}  
\usepackage{helvet} 
\usepackage{courier}  
\usepackage[hyphens]{url}  
\usepackage{graphicx} 
\usepackage{natbib}  
\usepackage{caption} 
\usepackage[utf8]{inputenc} 
\usepackage{url}            
\usepackage{booktabs}       
\usepackage{amsfonts}       
\usepackage{nicefrac}       
\usepackage{microtype}      
\usepackage{amsmath}
\usepackage{graphicx}
\usepackage{textcomp}
\usepackage{caption}
\usepackage{subcaption}
\usepackage{epigraph}
\usepackage[export]{adjustbox}
\usepackage{booktabs} 
\usepackage{algorithm}
\usepackage{subcaption}
\usepackage[noend]{algpseudocode}
\usepackage[inline]{enumitem}
\usepackage[title]{appendix}
\usepackage{xifthen}
\usepackage{tabularx}
\usepackage{xspace}
\usepackage{multirow}
\usepackage{booktabs}

\usepackage{makecell}
\usepackage{arydshln}

\usepackage{soul}

\algnewcommand\algorithmicforeach{\textbf{for each}}
\algdef{S}[FOR]{ForEach}[1]{\algorithmicforeach\ #1\ \algorithmicdo}

\newcommand{\sudoku}[1][]{\ifthenelse{\isempty{#1}}{$\mathrm{sudoku}$}{$\mathrm{sudoku}(#1)$}}
\newcommand{\queens}[1][]{\ifthenelse{\isempty{#1}}{$\mathrm{n\mbox{-}queens}$}{$\mathrm{n\mbox{-}queens}(#1)$}}
\newcommand{\sha}[1][]{\ifthenelse{\isempty{#1}}{$\mathrm{sha\textnormal{-}1}$}{$\mathrm{sha\textnormal{-}1}(#1)$}}
\newcommand{\cell}[1][]{\ifthenelse{\isempty{#1}}{$\mathrm{cell}$}{$\mathrm{cell}(#1)$}}
\newcommand{\island}[1][]{\ifthenelse{\isempty{#1}}{$\mathrm{island}$}{$\mathrm{island}(#1)$}}
\newcommand{\grid}[1][]{\ifthenelse{\isempty{#1}}{$\mathrm{grid\_wrld}$}{$\mathrm{grid\_wrld}(#1)$}}
\newcommand{\arith}[1][]{\ifthenelse{\isempty{#1}}{$\mathrm{bv\_expr}$}{$\mathrm{bv\_expr}(#1)$}}
\newcommand{\itarith}[1][]{\ifthenelse{\isempty{#1}}{$\mathrm{it\_expr}$}{$\mathrm{it\_expr}(#1)$}}
\newcommand{\sharpsat}{\texttt{SharpSAT}\xspace}
\newcommand{\neurosharp}{\texttt{Neuro\#}\xspace}
\newcommand{\ganak}{\textsc{Ganak}\xspace}

\algnewcommand{\IIf}[1]{\State\algorithmicif\ #1}
\algnewcommand{\EndIIf}{\unskip\ \algorithmicend\ \algorithmicif}

\newcolumntype{R}[2]{%
    >{\adjustbox{angle=#1,lap=\width-(#2)}\bgroup}%
    l%
    <{\egroup}%
}
\newcommand{\rot}[2]{\multicolumn{1}{R{#2}{2em}}{#1}}

\newcommand\blfootnote[1]{%
\begingroup
\renewcommand\thefootnote{}\footnote{#1}%
\addtocounter{footnote}{-1}%
\endgroup
}

\usepackage{nccfoots}

\title{Learning Branching Heuristics for Propositional Model Counting}

\author {
    Pashootan Vaezipoor,\textsuperscript{\rm 1*}\blfootnote{Equal Contribution}
    Gil Lederman,\textsuperscript{\rm 2*}
    Yuhuai Wu,\textsuperscript{\rm 1,\rm 3}
    Chris Maddison,\textsuperscript{\rm 1,\rm 3}\\
    Roger B. Grosse,\textsuperscript{\rm 1,\rm 3}
    Sanjit A. Seshia,\textsuperscript{\rm 2}
    Fahiem Bacchus\textsuperscript{\rm 1}\\
}
\affiliations {
    \textsuperscript{\rm 1} University of Toronto\\
    \textsuperscript{\rm 2} UC Berkeley\\
    \textsuperscript{\rm 3} Vector Institute\\
    \{pashootan, ywu, cmaddis, rgrosse, fbacchus\}@cs.toronto.edu\\ \{gilled, sseshia\}@eecs.berkeley.edu
}

\begin{document}

\maketitle

\begin{abstract}


Propositional model counting, or \#SAT, is the problem of computing
the number of satisfying assignments of a Boolean formula. Many
problems from different application areas, including many discrete probabilistic inference problems, can be translated into model counting problems to be 
solved by \#SAT solvers. Exact \#SAT solvers, however, 
are often not scalable to industrial size instances. In this paper, we 
present \neurosharp, an approach for learning branching heuristics to improve 
the performance of exact \#SAT 
solvers on instances from a given family of problems. 
We experimentally show that our method reduces the step count
on similarly distributed held-out instances and generalizes to much larger instances from the same problem family. It is able to achieve these results on a number of different problem families having very different structures.
In addition to step count improvements, \neurosharp can also achieve orders of magnitude wall-clock speedups over the vanilla solver on larger instances in some problem families, despite the runtime overhead of querying the model.


\end{abstract}
\section{Introduction}

Propositional model counting is the problem of counting the number of
satisfying solutions to a Boolean formula \cite{DBLP:series/faia/GomesSS09}. When the Boolean formula
is expressed in conjunctive normal form (CNF), this problem is known as
the \#SAT problem. \#SAT is a
\#P-complete problem, and by Toda's theorem
\cite{DBLP:journals/siamcomp/Toda91} any problem in the
polynomial-time hierarchy (PH) can be solved by a polynomial number of
calls to a \#SAT oracle. This means that effective \#SAT solvers, if they could be
developed, have the potential to help solve problems whose complexity lies beyond NP,
from a range of applications.
The tremendous practical successes achieved 
by encoding problems to SAT and using modern SAT solvers \cite{DBLP:conf/cie/Marques-Silva18} 
demonstrate the potential of such an approach.


Modern exact \#SAT solvers are based on the DPLL algorithm \cite{DBLP:journals/cacm/DavisLL62} and have been successfully
applied to solve certain problems, e.g., inference in Bayes Nets~\cite{DBLP:conf/uai/BacchusDP03,DBLP:journals/jair/0002PB11,DBLP:conf/aaai/SangBK05,DBLP:journals/jair/DomshlakH07} and bounded-length probabilistic planning~\cite{DBLP:conf/aips/DomshlakH06};
however, many applications remain out of reach of current solvers. For example, in problems such as inference in Markov Chains, which have a temporal structure, exact model counters are still generally inferior to earlier methods such as Binary Decision Diagrams (BDDs).
In this paper we show that machine learning methods can be used to
greatly enhance the performance of exact \#SAT solvers.

In particular, we learn problem family specific branching heuristics for the 2012 version of the DPLL-based \#SAT solver \sharpsat \cite{thurley2006sharpsat} which uses a state-of-the-art search procedure. We cast the problem as a Markov Decision Process (MDP) in which the task is to select the best literal to branch on next. We use a Graph Neural Network (GNN) \cite{ScarselliGTHM09} to represent the particular component of the residual formula the solver is currently working on. 
The model is trained end-to-end via Evolution Strategies (ES), with the objective of minimizing the mean number of branching decisions required to solve instances from \emph{a given distribution of problems}. In other words, given a training set of instances drawn from a problem distribution, the aim is to automatically tailor the solver's branching decisions for better performance on unseen problems of that distribution. 






We found that our technique, which we call \neurosharp, can generalize not only to unseen problem instances of similar size but also to much larger instances than those seen at training time. Furthermore, despite \neurosharp's considerable runtime overhead from querying the learnt model, on some problem domains \neurosharp can achieve \emph{orders-of-magnitude improvements} in
the solver's \emph{wall-clock} runtime. This is quite remarkable in the context
of prior related work~\cite{YolcuP19,selsam2019neurocore,DBLP:conf/icml/BalcanDSV18,DBLP:conf/nips/GasseCFC019,DBLP:conf/aaai/KhalilBSND16,hansknecht2018cuts,LedermanRSL20},
where using ML to improve combinatorial solvers has at best yielded
modest wall-clock time improvements (less than a factor of
two), and positions this line of research as a viable path towards improving the practicality of exact model counters.\footnote{Our code and the extended version of the paper (with the appendix) are available at: \texttt{github.com/NeuroSharp}.}

\subsection{Related Work}\label{sec:related}
The first successful application of machine learning to propositional satisfiability solvers was the \emph{portfolio-based} SAT solver \texttt{SATZilla} \cite{xu2008satzilla}. Equipped with a set of standard SAT solvers, a classifier was trained offline to map a given SAT instance to the solver best suited to solve it. 

Recent work has been directed along two paths: \emph{heuristic improvement} \cite{selsam2019neurocore, kurin2019improving, LedermanRSL20, YolcuP19}, and purely ML-based solvers \cite{selsam2018learning, amizadeh2018learning}. In the former, 
a model is trained to replace a heuristic in a standard solver, thus the model is embedded as a module within the solver's framework and guides the search process. 
In the latter, the aim is to train a model that acts as a \emph{stand-alone} ``neural'' solver. These neural solvers are inherently stochastic and often \emph{incomplete}, meaning that they can only provide an estimate of the satisfiability of a given instance. This is often undesirable in applications of SAT solvers where an exact answer is required, e.g., in formal verification. In terms of functionality, our work is analogous to the first group, in that we aim at improving the branching heuristics of a standard solver. 
More concretely, our work is similar to \cite{YolcuP19}, who used Reinforcement Learning (RL) and GNNs to learn variable selection heuristics for the \emph{local search-based} SAT solver \texttt{WalkSAT} \cite{selman1993local}. Local search cannot be used to solve exact \#SAT, and empirically \cite{YolcuP19} obtained only modest improvements on much smaller problems. Our method is also related to \cite{LedermanRSL20} and \cite{DBLP:conf/nips/GasseCFC019}, where similar techniques were used in solving quantified Boolean formulas and mixed integer programs, respectively. 

Recently, \citet{DBLP:conf/aaai/AbboudCL20} trained GNNs as a stand-alone approximate solver for Weighted DNF Model Counting (\#DNF). However, approximating \#DNF is a much easier problem: it has a fully polynomial randomized approximation scheme \cite{DBLP:journals/jal/KarpLM89}. So the generalization to larger problem instances demonstrated in that paper is not comparable to the scaling on exact \#SAT our approach achieves. 



\section{Background}\label{sec:background}
\paragraph{\#SAT}
\label{sec:notations}
A propositional Boolean formula consists of a set of propositional
(true/false) variables composed by applying the standard operators ``and''
($\land$), ``or'' ($\lor$) and ``not'' ($\lnot$). A \emph{literal} is any
variable $v$ or its negation $\lnot v$. A \emph{clause} is a
disjunction of literals $\bigvee_{i=1}^n l_i$. A clause is 
a \emph{unit clause} if it contains only one literal.  Finally, a
Boolean formula is in \emph{Conjunctive Normal Form} (CNF) if it is a
conjunction of clauses. We denote the set of literals and clauses of a CNF 
formula $\phi$ by $\mathcal{L}(\phi)$ and $\mathcal{C}(\phi)$, respectively. We assume that all formulas are in CNF.

A \emph{truth assignment} for any formula $\phi$ is a
mapping of its variables to $\{0, 1\}$ (\textbf{false}/\textbf{true}). Thus there are
$2^n$ different truth assignments when $\phi$ has $n$ variables. A 
truth assignment $\pi$ satisfies a literal $\ell$ when $\ell$ is the
variable $v$ and $\pi(v)=1$ or when $\ell = \lnot v$ and $\pi(v) =
0$. It satisfies a clause when at least one of its
literals is satisfied. A CNF formula $\phi$ is satisfied when all of its
clauses are satisfied under $\pi$ in which case we call $\pi$ a 
\emph{satisfying assignment} for $\phi$. 

The \#SAT problem for $\phi$ is to compute the number of
satisfying assignments. If $\ell$ is a unit clause of $\phi$ then all of
$\phi$'s satisfying assignments must make $\ell$ true. If another clause
$c'= \lnot \ell \lor \ell'$ is in $\phi$, then every satisfying assignment
must also make $\ell'$ true since $\lnot \ell \in c'$ must be false. This
process of finding all literals whose truth value is forced by unit
clauses is called \emph{Unit Propagation} (UP) and is used in all SAT and
\#SAT solvers. 
Such solvers traverse the search tree by employing a branching heuristic.
This heuristic selects an unforced variable and branches on it by setting it to \textbf{true} or \textbf{false}.
When a literal $\ell$ is set to \textbf{true} the formula
$\phi$ can be reduced by finding all forced literals using UP (this
includes $\ell$), removing all clauses containing a true literal, and
finally removing all false literals from all clauses. The resulting formula
is denoted by $\textrm{UP}(\phi, \ell)$.

Two sets of clauses are called \emph{disjoint} if they share no variables. A component $C\subset \mathcal{C} (\phi)$ is a subset of $\phi$'s clauses that is disjoint from its complement $\mathcal{C}(\phi) - C$. A formula $\phi$ can be
efficiently broken up into a maximal number of disjoint components
$C_1$, \ldots, $C_k$. Although most formulas initially consist of only one
component, as variables are set by branching decisions and clauses are
removed, the reduced formulas will often break up into multiple
components. Components are important for improving the efficiency of
\#SAT solving as each component can be solved separately and their
counts multiplied: $\textrm{COUNT}(\phi) = \prod_{i=1}^{k} \textrm{COUNT}(C_i)$. In contrast, solving the formula as a monolith takes $2^{\Theta(n)}$
where $n$ is the number of variables in the input formula, and so is not efficient
for large $n$.

\begin{figure}[ht]
\centering
\includegraphics[width=0.3\textwidth]{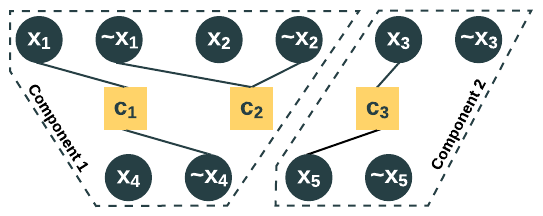}
\caption{An example Literal-Clause Incidence Graph (LIG) for formula:
  $(x_1\vee \neg x_4)\wedge(\neg x_1\vee \neg x_2)\wedge(x_3 \vee
  x_5)$.}
\label{fig:incident}
\begin{minipage}{0.43\textwidth}
\algrenewcommand\algorithmicindent{1.0em}
\begin{algorithm}[H]
\begin{algorithmic}[1]
\Function{\#DPLLCache}{$\phi$}
  \If {\textproc{inCache}($\phi$)}
    \State \Return \textproc{cacheLookUp}($\phi$)
  \EndIf
  \State Pick a literal $\ell \in \mathcal{L}(\phi)$ \label{alg:line:branch_select}
  \State \#$\ell$ = \textproc{CountSide}($\phi$, $\ell$)
  \State \#$\lnot \ell$ = \textproc{CountSide}($\phi$, $\lnot \ell$)
  \State \textproc{addToCache}($\phi$, \#$\ell$ \;+\; \#$\lnot \ell$)\label{alg:line:cache_value}
  \State \Return \#$\ell$ \;+\; \#$\lnot \ell$
\EndFunction
\Statex
\Function{CountSide}{$\phi$, $\ell$}
  \State $\phi_\ell = \mbox{}$ \textrm{UP}($\phi$, $\ell$)
  \If {$\phi_\ell$ contains an empty clause}
     \State \Return 0
  \EndIf
  \If {$\phi_\ell$ contains no clauses}
     \State $k$ = \# of unset variables
     \State \Return $2^{k}$
  \EndIf
  \State $K = \textproc{findComponents}(\phi_\ell)$ 
  \State \Return $\prod_{\kappa\in K} \textproc{\#DPLLCache}(\kappa)$\label{alg:line:prod}

\EndFunction
\end{algorithmic}
\caption{Component Caching DPLL}\label{alg:sharp_dpll}
\end{algorithm}
\end{minipage}
\end{figure}

A formula $\phi$ or component $C_i$ can be represented by a \emph{literal-clause incidence
  graph} (LIG).
This graph contains a node for every clause and every literal of $\phi$ (i.e., $v$ and $\neg v$ for every variable $v$ of $\phi$). An edge connects
a clause node $n_c$ and a literal node $n_\ell$ if and only if $\ell\in c$. Note that if $C_i$ is a component of $\phi$, then $C_i$'s LIG will be a disconnected sub-graph of $\phi$'s LIG (Figure~\ref{fig:incident}).

Both exact
\cite{thurley2006sharpsat,DBLP:conf/sat/SangBBKP04,DBLP:conf/ijcai/OztokD15},
approximate \cite{DBLP:conf/aaai/ChakrabortyFMSV14,DBLP:journals/corr/abs-2004-14692}, and probabilistically correct  \cite{DBLP:conf/ijcai/SharmaRSM19} model counters have been developed. In this paper, we focus on exact model counting 
using the \sharpsat{} solver
\cite{thurley2006sharpsat}. \sharpsat{} and other modern exact \#SAT
solvers are based on DPLL \cite{DBLP:journals/cacm/DavisLL62} augmented with \emph{clause
learning} and \emph{component caching}
\cite{DBLP:conf/focs/BacchusDP03,DBLP:journals/jair/BacchusDP09}. 
A simplified version of the algorithm 
with the clause learning parts omitted is given in 
Algorithm~\ref{alg:sharp_dpll}. 

The \textproc{\#DPLLCache} algorithm works on one component at a time. If that component's
model count has already been cached it returns the cached
value. Otherwise it selects a literal to branch on
(line~\ref{alg:line:branch_select}) and computes the model count under
each value of this literal by calling \textproc{CountSide()}. The sum of these two counts is the model
count of the passed component $\phi$, and so is stored in the cache
(line~\ref{alg:line:cache_value}). The \textproc{CountSide} function
first unit propagates the input literal. If an empty clause is found, then the
current formula $\phi_\ell$ is unsatisfiable and has zero
models. Otherwise, $\phi_\ell$ is divided into its sub-components which are
independently solved and the product of their model counts is
returned. Critical to the performance of the algorithm is the choice
of which literal from the current formula $\phi$ to branch on. This
choice affects  the efficiency of clause learning and the effectiveness
of component generation and caching lookup success. \sharpsat uses the VSADS heuristic
\cite{DBLP:conf/sat/SangBK05} which is a linear combination of a
  heuristic aimed at making clause learning effective (VSIDS) and a
  count of the number of times a variable appears in the current
  formula.

\paragraph{Graph Neural Networks}
\emph{Graph Neural Networks} (GNNs) are a class of neural networks used for representation learning over graphs \cite{gnn_gori, ScarselliGTHM09}. Utilizing a neighbourhood aggregation (or message passing) scheme, GNNs map the nodes of the input graph to a vector space. 
Let $G=(V, E)$ be an undirected graph with node feature vectors $h^{(0)}_v$ for each node $v\in V$. GNNs use the graph structure and the node features to learn an embedding vector $h_v$ for every node. This is done through iterative applications of a neighbourhood aggregation function. In each iteration $k$, the embedding of a node $h^{(k)}_v$ is updated by aggregating the embeddings of its neighbours from iteration $k-1$ and passing the result through a nonlinear aggregation function $A$ parameterized by $W^{(k)}$:
\begin{equation}
    h^{(k)}_v = \textsc{A} \Big( h^{(k - 1)}_v , \sum_{u\in\mathcal{N}(v)}h^{(k-1)}_u; W^{(k)} \Big),
\end{equation}
where $\mathcal{N}(v) = \{u| u \in V \wedge (v, u) \in E\}$. After $K$ iterations, $h^{(K)}_v$ is extracted as the final node embedding $h_v$ for node $v$. Through this scheme, $v$'s node embedding at step $k$ incorporates the structural information of all its $k$-hop neighbours. 

\paragraph{Evolution Strategies}
\label{sec:es}
\emph{Evolution Strategies} (ES) are a class of zeroth order black-box optimization algorithms~\cite{DBLP:journals/nc/BeyerS02,DBLP:journals/jmlr/WierstraSGSPS14}. Inspired by natural evolution, a population of parameter vectors (genomes) is perturbed (mutated) at every iteration, giving birth to a new generation. The resulting offspring are then evaluated by a predefined fitness function. Those offspring with higher fitness score will be selected for producing the next generation.

We adopt a version of ES that has shown to achieve great success in the standard RL benchmarks  \cite{DBLP:journals/corr/SalimansHCS17}: Let $f:\Theta\to\mathbb{R}$ denote the fitness function for a parameter space $\Theta$, e.g., in an RL environment, $f$ computes the stochastic episodic reward of a policy $\pi_{\theta}$. To produce the new generation of parameters of size $n$, \cite{DBLP:journals/corr/SalimansHCS17} uses an additive Gaussian noise with standard deviation $\sigma$ to perturb the current generation: $\theta^{(i)}_{t+1}=\theta_t+\sigma\epsilon^{(i)}$, where $\epsilon^{(i)}\sim\mathcal{N}(0, I)$. We then evaluate every new generation with fitness function $f(\theta^{(i)}_{t+1})$ for all $i\in[1,\dots,n]$. The update rule of the parameter is as follows,
\begin{align*}
    \theta_{t+1} &=\theta_t + \eta\nabla_{\theta}\mathbb{E}_{\theta\sim\mathcal{N}(\theta_t,\sigma^2I) }[f(\theta)]\\
    &\approx\theta_{t} + \eta\frac{1}{n\sigma}\sum_{i}^n f(\theta^{(i)}_{t+1})\epsilon^{(i)},
\end{align*}
where $\eta$ is the learning rate. The update rule is intuitive: each perturbation $\epsilon^{(i)}$ is weighted by the fitness of the corresponding offspring $\theta^{(i)}_{t+1}$. We follow the rank-normalization and mirror sampling techniques of~\cite{DBLP:journals/corr/SalimansHCS17} to scale the reward function and reduce the variance of the gradient, respectively.
\section{Method}\label{sec:architecture}

We formalize the problem of learning the branching heuristic for $\textproc{\#DPLLCache}$ as a \emph{Markov Decision Process} (MDP). In our setting, the environment is \texttt{SharpSAT}, which is deterministic except for the initial state, where an instance (CNF formula) is chosen randomly from a given distribution. A time step $t$ is equivalent to an invocation of the branching heuristic by the solver (Algorithm~\ref{alg:sharp_dpll}: line \ref{alg:line:branch_select}). At time step $t$ the agent observes state $s_t$, consisting of the component $\phi_t$ that the solver is operating on, and performs an action from the action space $\mathcal{A}_t = \{l|l\in\mathcal{L}(\phi_t)\}$. The objective function is to reduce the number of decisions the solver makes, while solving the counting problem. 
In detail, the reward function is defined by:
\begin{equation*}
  R(s) =
    \begin{cases}
      1 & \textit{\parbox{5cm}{if $s$ is a terminal state with \emph{``instance solved''} status\smallskip}}\\
      -r_{penalty} & \textit{otherwise}
    \end{cases}       
\end{equation*}
If not finished, episodes are aborted after a predefined max number of steps, without receiving the termination reward.

\paragraph{Training with Evolution Strategies} With the objective defined, we observe that for our task, the potential action space as well as the horizon of the episode can be quite large (up to 20,000 and 1,000, respectively).
As~\citet{vemula2019contrasting} show, the exploration complexity of an action-space exploration RL algorithm (e.g, Q-Learning, Policy Gradient) increases with the size of the action space and the problem horizon. On the other hand, a parameter-space exploration algorithm like ES is independent of these two factors. Therefore, we choose to use a version of ES proposed by~\citet{DBLP:journals/corr/SalimansHCS17} for optimizing our agent. 

\paragraph{SharpSAT Components as GNNs}\label{subsec:model}
As the task for the neural network agent is to pick a literal $l$ from the component $\phi$, we opt for a LIG representation of the component (see Section \ref{sec:background}) and we use GNNs to compute a literal selection heuristic based on that representation. 
In detail, given the LIG $G = (V, E)$ of a component $\phi$, we denote the set of clause nodes as $C\subset V$, and the set of literal nodes as $L\subset V$, $V=C\cup L$. 
The initial vector representation is denoted by $h^{(0)}_c$ for each clause $c\in C$ and $h^{(0)}_l$ for each literal $l\in L$, both of which are learnable model parameters. 
We run the following message passing steps iteratively:
\begin{align*}
   \textrm{- Literal to }&\textrm{Clause (L2C):}\\
   h^{(k+1)}_c &= \mathcal{A} \Big( h^{(k)}_c, \sum_{l\in c}[h^{(k)}_l, h^{(k)}_{\bar{l}}]; W_C^{(k)}\Big), \quad \forall c\in C,\\
   \textrm{- Clause to }&\textrm{Literal (C2L):}\\
   h^{(k+1)}_l &= \mathcal{A} \Big( h^{(k)}_l, \sum_{c, l\in c}h^{(k)}_c; W_L^{(k)} \Big), \quad \forall l\in L,
\end{align*}
where $\mathcal{A}$ is a nonlinear aggregation function, parameterized by $W_C^{(k)}$ for clause aggregation and $W_L^{(k)}$ for literal aggregation at the $k^{th}$ iteration. Following~\citet{selsam2018learning, LedermanRSL20}, to ensure negation invariance (i.e.~that the graph representation is invariant under literal negation), we concatenate the literal representations corresponding to the same variable $h^{(k)}_l, h^{(k)}_{\bar{l}}$ when running L2C message passing.
 After $K$ iterations, we obtain a $d$-dimensional vector representation for every literal in the graph. We pass each representation through a policy network, a Multi-Layer Perceptron (MLP), to obtain a score, and we choose the literal with the highest score. Recently, \citet{xu2018powerful} developed a GNN architecture named \emph{Graph Isomorphism Network} (GIN), and proved that it achieves maximum expressiveness among the class of GNNs. We hence choose GIN for the parameterization of $\mathcal{A}$. Specifically, $\mathcal{A}(x,y;W)=\textsc{MLP}((1+\epsilon)x+y;W)$,
 where $\epsilon$ is a hyperparameter. 

\paragraph{Sequential Semantics} \label{subsec:semantic}

Many problems, such as dynamical systems and bounded model checking, are iterative in nature, with a distinct temporal dimension to them. In the original problem domain, there is often a state that is evolved through time via repeated applications of a state transition function. A structured CNF encoding of such problems usually maps every state $s_t$ to a set of variables, and adds sets of clauses to represent the dynamical constraints between every transition $(s_t,s_{t+1})$. Normally, all temporal information is lost in reduction to CNF. However, with a learning-based approach, the time-step feature from the original problem can be readily incorporated as an additional input to the network, effectively annotating each variable with its time-step. In our experiments, we represented time by appending to each literal embedding a scalar value (representing the normalized time-step $t$) before passing it through the output MLP. We perform an ablation study to investigate the impact of this additional feature in Section~\ref{sec:exp}.

\paragraph{Engineering Trade-offs and Constraints} \label{subsec:tradeoffs}
Directly training on challenging \#SAT instances of enormous size is computationally infeasible. We tackle this issue by training \neurosharp on small instances of a problem (fast rollouts) and relying on generalization to solve the more challenging instances from the same problem domain. Thus the main requirement is the availability of the generative process that lets us sample problems with desired level of difficulty. Access to such generative process is not an unreasonable assumption in industry and research.

Although reducing the number of branching steps is itself interesting, to beat \sharpsat{} in wall-clock time, \neurosharp{}'s lead needs to be wide enough to justify the imposed overhead of querying the GNN. Since the time-per-step ratio is relatively constant, for the method to be effective it is desirable that the step count reduction be \emph{superlinear}, meaning it becomes more effective compared to the vanilla heuristic the larger the problem becomes.

\section{Data Generation}\label{sec:data}

\begin{figure}[tb]
   \centering
\begin{tabular}{ll}
\includegraphics[width=7cm]{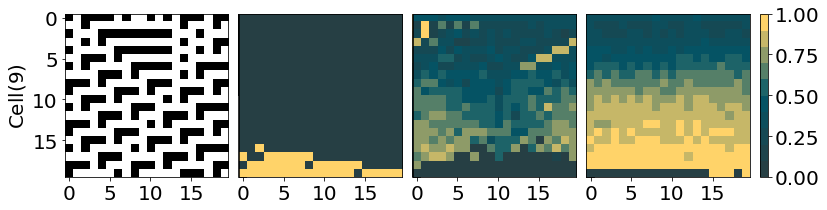}\\
\includegraphics[width=7cm]{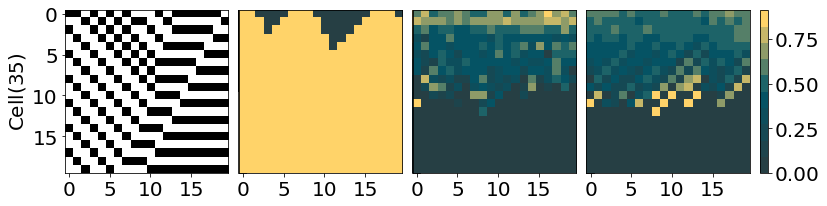}\\
\includegraphics[width=7cm]{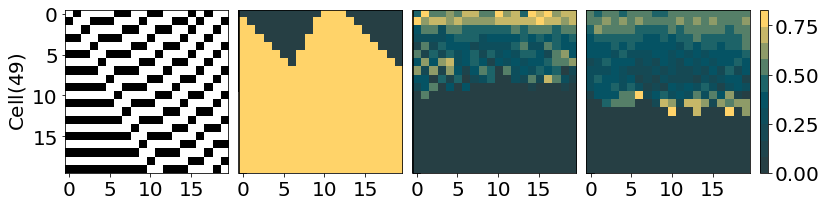}\\
\small{\hspace{3em} (a) \hspace{3em} (b) \hspace{3em} (c)\hspace{3.4em} (d)}
\end{tabular}
    \caption{Contrary to \sharpsat, \neurosharp branches earlier on variables of the bottom rows. (a) Evolution of a bit-vector through repeated applications of Cellular Automata rules. The result of applying the rule at each iteration is placed under the previous bit-vector, creating a two-dimensional, top-down representation of the system's evolution; (b) The initial formula simplification on a \emph{single} formula. Yellow indicates the regions of the formula that this process prunes; (c~\&~d) Variable selection ordering by \sharpsat{} and \neurosharp averaged over the entire dataset. Lighter colours show that the corresponding variable is selected earlier on average.}
    \label{fig:cells}
\end{figure}

Our goal was to evaluate our method on more structured and much larger instances than the small random instances typically used in other related works \cite{YolcuP19, selsam2018learning, kurin2019improving}. To that end, we searched SAT and planning benchmarks for problems whose generative processes were publicly available or feasible to implement. To test the versatility of our method, we made sure that these problems cover a diverse set of domains: \sudoku[], blocked \queens[], \cell[] \emph{(combinatorial)}; \sha[]  preimage attack \emph{(cryptography)}; \island[], \grid[] (\emph{planning}), \arith[], \itarith[] \emph{(circuits)}. For brevity, we explain two of the problems that we will use in later sections to discuss the behaviour of our trained model and provide a more detailed description of the other datasets and their generation process in Appendix~\ref{appendix:dataset}:

\begin{itemize}




\item \cell[R,n,r]: Elementary (i.e., one-dimensional, binary) Cellular Automata are simple systems of computation where the cells of an $n$-bit binary state vector are progressed through time by repeated applications of a rule $R$ (seen as a function on the state space). 
Figure \ref{fig:cells}a shows the evolution grid of rules 9, 35 and 49 for 20 iterations.

Reversing Elementary Cellular Automata: 
Given a randomly sampled state $T$, compute the number of initial states $I$ that would lead to that terminal state $T$ in $r$ applications of $R$, i.e.,  $\big\vert\{I: R^r(I)=T\}\big\vert$. The entire $r$-step evolution grid is encoded by mapping each cell to a Boolean variable ($n\times r$ in total). The clauses impose the constraints between cells of consecutive rows as given by $R$. The variables corresponding to $T$ (last row of the evolution grid) are assigned as unit clauses. This problem was taken from SATCOMP 2018 \cite{heule2018proceedings}.

    
\item \grid[s,t]: This bounded horizon planning problem from \cite{vazquez-neurips18,gridencoding} is based on encoding a grid world with different types of squares (e.g., lava, water, recharge), and a formal specification such as \emph{``Do not recharge while wet''} or \emph{``avoid lava''}. 
We randomly sample a grid world of size $s$ and a starting position $I$ for an agent. 
We encode to CNF the problem of counting the number of trajectories of length $t$ beginning from $I$ that always avoid lava. 

\end{itemize}

\section{Experiments}\label{sec:exp}

\begin{table}[t]
  \centering
  \scriptsize
  \begin{tabular}{*{6}l@{\hskip 0in}l}
\toprule
\centering \textbf{Dataset} &   \rot{\# vars}{15}   &   \rot{\# clauses}{15}   & \rot{\texttt{Random}}{15} & \rot{\sharpsat}{15} & \rot{\neurosharp}{15}\\ 
\cmidrule(lr){2-6}
\sudoku[9,25]         &  182   &  3k  & 338 & 220 & \textbf{195(1.1x)}\\
\queens[10,20]             &  100   &  1.5k  & 981 & 466 & \textbf{261(1.7x)} \\
\sha[28]             &  3k   & 13.5k   & 2,911 & 52 & \textbf{24(2.1x)} \\
\island[2,5]            & 1k    &   34k & 155 & 86 & \textbf{30(1.8x)} \\
\cell[9,20,20]     &  210 &  1k  & 957 & 370 & \textbf{184(2.0x)} \\
\cell[35,128,110]  & 6k & 25k & 867 & 353 & \textbf{198(1.8x)}\\
\cell[49,128,110]       & 6k  & 25k & 843 & 338 & \textbf{206(1.6x)}\\
\grid[10,5]       & 329 & 967 & 220 & 195 & \textbf{66(3.0x)}\\
\arith[5,4,8]     &  90 & 220 & 1,316 & 328 & \textbf{205(1.6x)}\\
\itarith[2,2]             &  82   &  264  & 772 & 412 & \textbf{266(1.5x)}\\
\bottomrule
\end{tabular}
\captionof{table}{\neurosharp generalizes to unseen i.i.d. test problems often with a large margin compared to \sharpsat{}.} 
\label{table:res}
\end{table}

\begin{table}[tbh]
  \centering
  \scriptsize
  \begin{tabular}{*{6}l@{\hskip 0in}l}
\toprule

\textbf{Dataset} & \rot{\# vars}{15}   &   \rot{\# clauses}{15}   & \rot{\texttt{Random}}{15} & \rot{\sharpsat}{15} &\rot{\neurosharp}{15} \\
\cmidrule(lr){2-7}
\sudoku[16,105]        &  1k &   31k  & 7,654 &  2,373  & \textbf{2,300} & \textbf{(1.03x)}\\
\hline
\queens[12,20]        & 144 & 2.6k & 31,728 & 12,372 & \textbf{6,272} & \textbf{(1.9x)}\\
\hline
\sha[40]        & 5k  & 25k & 15k & 387 & \textbf{83} & \textbf{(4.6x)}\\
\hline
\island[2,8]            & 1.5k    &   73.5k & 1,335 & 193 & \textbf{46} & \textbf{(4.1x)} \\
\hline
\cell[9,40,40]    & 820 & 4k & 39,000 & 53,349 & \textbf{42,325} & \textbf{(1.2x)}\\
\hdashline[.4pt/1pt]
\cell[35,192,128] & 12k & 49k & 36,186 & 21,166 & \textbf{1,668} & \textbf{(12.5x)}\\
\cell[35,256,200] & 25k & 102k & 41,589 & 26,460 & \textbf{2,625} & \textbf{(10x)}\\
\cell[35,348,280] & 48k & 195k & 54,113 & 33,820 & \textbf{2,938} & \textbf{(11.5x)}\\
\hdashline[.4pt/1pt]
\cell[49,192,128]       & 12k & 49k & 35,957 & 24,992 & \textbf{1,829} & \textbf{(13.6x)}\\
\cell[49,256,200] & 25k & 102k & 47,341 & 30,817 & \textbf{2,276} & \textbf{(13.5x)}\\
\cell[49,348,280] & 48k & 195k & 53,779 & 37,345 & \textbf{2,671} & \textbf{(13.9x)}\\
\hline
\grid[10,10]      & 740 & 2k & 22,054 & 13,661 & \textbf{367} & \textbf{(37x)}\\
\grid[10,12]&  2k & 6k  & 100k$\le$ & 93,093 & \textbf{1,320} & \textbf{(71x)}\\
\grid[10,14]&  2k & 7k  & 100k$\le$ & 100k$\le$ & \textbf{2,234} & \textbf{(--)}\\ 
\grid[12,14]&  2k & 8k  & 100k$\le$ & 100k$\le$ & \textbf{2,782} & \textbf{(--)}\\
\hline
\arith[7,4,12]    & 187 & 474 & 35,229& 5,865 & \textbf{2,139} & \textbf{(2.7x)}\\
\hline
\itarith[2,4]        &  162 & 510 & 51,375 & 7,894 & \textbf{2,635} & \textbf{(3x)}\\
\bottomrule
\end{tabular}
\captionof{table}{\neurosharp generalizes to much larger problems than what it was trained on, sometimes achieving orders of magnitude improvements over \sharpsat{}.}
\label{table:res-upwards}
\end{table}

\begin{figure}[thb]
\centering
\begin{tabular}{cc}
\includegraphics[width=0.3\textwidth]{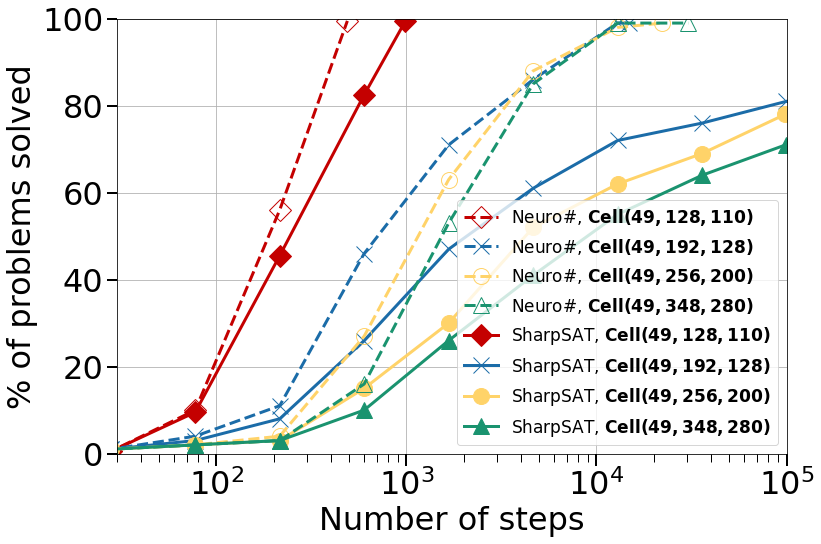}&
\raisebox{1.5\height}{\rotatebox{90}{(a) \cell}}\\
\includegraphics[width=0.3\textwidth]{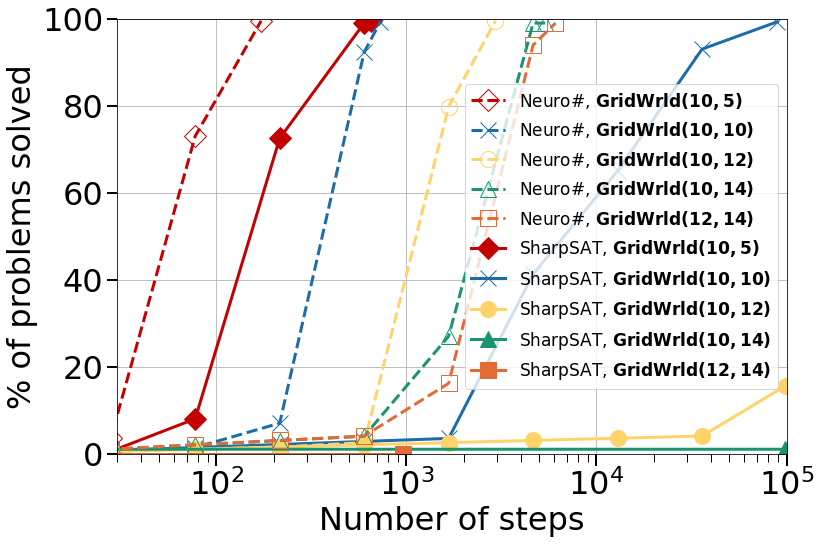}&
\raisebox{0.7\height}{\rotatebox{90}{(b) \grid}}\\
\end{tabular}
\captionof{figure}{\neurosharp generalizes well to larger problems. Compare the robustness of \neurosharp vs. \sharpsat{} as the problem sizes increase. Solid and dashed lines correspond to \sharpsat{} and \neurosharp, respectively. All episodes are capped at 100k steps.}
\label{fig:upward}
\end{figure}

To evaluate our method, we designed experiments to answer the following questions:
\begin{enumerate*}[label=\textbf{\arabic*)}]
\item \textbf{I.I.D. Generalization:} Can a model trained on instances from a given distribution generalize to unseen instances of the same distribution?
\item \textbf{Upward Generalization:} Can a model trained on small instances generalize to larger ones?
\item \textbf{Wall-Clock Improvement:} Can the model improve the runtime substantially?
\item \textbf{Interpretation:} 
Does the sequence of actions taken by the model exhibit any discernible pattern at the problem level?
\end{enumerate*} Our baseline in all comparisons is \sharpsat{}'s heuristic. Also, to make sure that our model's improvements are not trivially attainable without training we tested a \texttt{Random} policy that simply chooses a literal uniformly at random. We also studied the impact of the trained model on a variety of solver-specific quality metrics (e.g., cache-hit rate, \ldots), the results of which are in Appendix~\ref{appendix:solver_analysis}.



The \grid{}, being a problem of an iterative nature (i.e., steps in the planning problem), was a natural candidate for testing our hypothesis regarding the effect of adding the ``time'' feature of Section \ref{subsec:semantic}, so we report the results for \grid{} with that feature included and later in this section we perform an ablation study on that feature.

\paragraph{Experimental Setup}
For each dataset, we sampled 1,800 instances for training and 200 for testing. We trained for 1000 ES iterations. At each iteration, we sampled 8 formulas and 48 perturbations ($\sigma=0.02$). With mirror sampling, we obtained in total $96=48\times2$ perturbations. For each perturbation, we ran the agent on the 8 formulas (in parallel), to a total of $768=96\times8$ episodes per parameter update. All episodes, unless otherwise mentioned, were capped at 1k steps during training and 100k during testing. The agent received a negative reward of $r_{penalty} = 10^{-4}$ at each step. We used the Adam optimizer \cite{kingma2014adam} with default hyperparameters, a learning rate of $\eta=0.01$ and a weight decay of 0.005.

GNN messages were implemented by an MLP with ReLU non-linearity. The size of literal and clause embeddings were $32$ and the dimensionality of C2L \emph{(resp. L2C)} messages was $32\times32\times32$ \emph{(resp. $64\times32\times32$)}.
We used $T=2$ message passing iterations and final literal embeddings were passed through the MLP policy network of dimensions $32\times256\times64\times1$ to get the final score. When using the extra ``time'' feature, the first dimension of the decision layer was $33$ instead of 32.
The initial ($T=0$) embeddings of both literals and clauses were trainable model parameters.


\subsection{Results}\label{sec:results}
\paragraph{I.I.D. Generalization}
\label{sec:iid}
Table \ref{table:res} summarizes the results of the i.i.d.~generalization over the problem domains of Section \ref{sec:data}. We report the average number of branching steps on the test set. \neurosharp outperformed the baseline across all datasets. Most notably, on \grid{}, it reduced the number of branching steps by a factor of 3.0 
and on \cell{}, by an average factor of 1.8 over the three cellular rules. 



\paragraph{Upward Generalization}\label{sub:upwards}
We created instances of larger sizes (up to an order of magnitude more clauses and variables) for each of the datasets in Section \ref{sec:data}. We took the models trained from the previous i.i.d. setting and directly evaluated on these larger instances without further training. The evaluation results in Table \ref{table:res-upwards} show that \neurosharp generalized to the larger instances across all datasets and in almost all of them achieved substantial gains compared to the baseline as we increased the instance sizes.
Figure \ref{fig:upward} shows this effect for multiple sizes of \cell[49] and \grid{} by plotting the percentage of the problems solved within a number of steps. 
The superlinear gaps get more pronounced once we remove the cap of $10^{5}$ steps, i.e.,~let the episodes run to completion. In that case, on \grid[10, 12], \neurosharp took an average of 1,320 branching decisions, whereas \sharpsat{} took 809,408 (613x improvement).

\begin{figure}[tb]
\centering
\begin{tabular}{cc}
\includegraphics[width=0.45\columnwidth]{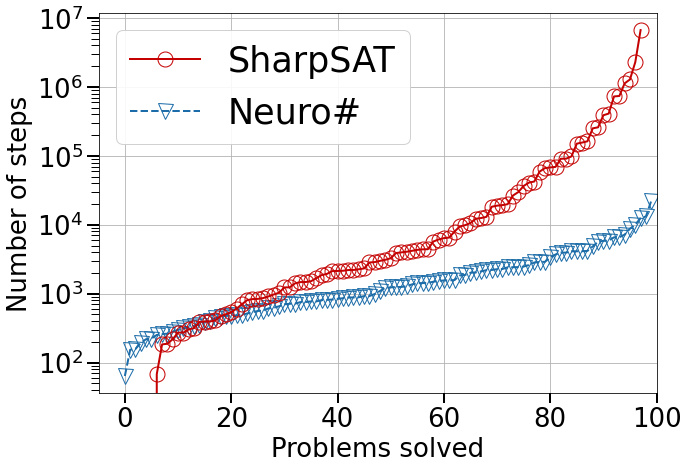}&
\includegraphics[width=0.45\columnwidth]{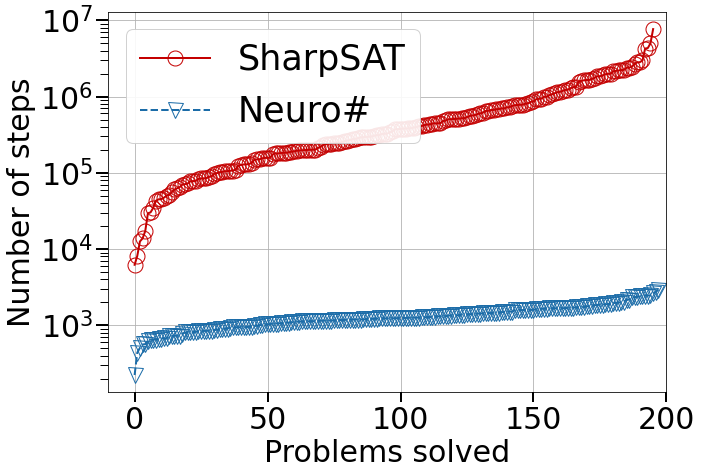}\\
\includegraphics[width=0.45\columnwidth]{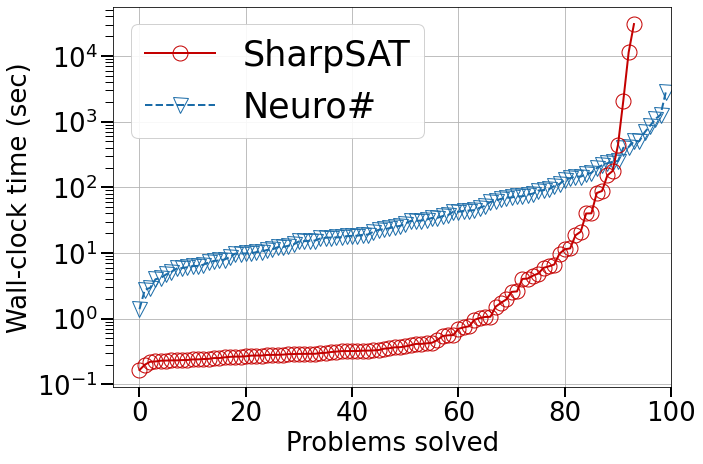}&
\includegraphics[width=0.45\columnwidth]{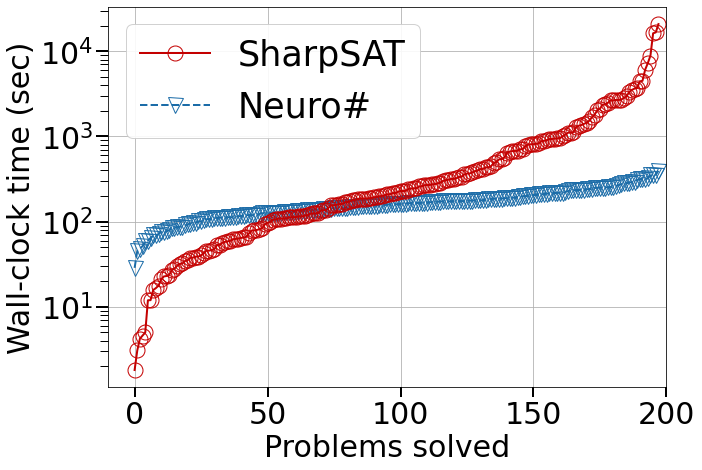}\\
(a) \cell[49,256,200] & (b) \grid[10,12]
\end{tabular}
\caption{Cactus plots comparing \neurosharp to \sharpsat on \cell{} and \grid{}. Lower and to the right is better: for any point $t$ on the $y$ axis, the plot shows the number of benchmark problems that are individually solvable by the solver, within $t$ steps (top) and seconds (bottom). }
\label{fig:wall_clock}
\end{figure}


\paragraph{Wall-Clock Improvement} 
Given the scale of improvements on the upward generalization benchmark, in particular \cell[49] and \grid, we measured the runtime of \neurosharp{} vs. \sharpsat on those datasets (Figure \ref{fig:wall_clock}). On both problems we observe that as \neurosharp{} widens the gap in the number of steps, it manages to beat \sharpsat{} in wall-clock. Note that the query overhead could still be greatly reduced in our implementation through GPU utilization, loading the model in the solver's code in C++ instead of making out-of-process calls to Python, etc.

\subsection{Model Interpretation} 
\begin{figure}[tb]
\centering
\includegraphics[width=0.38\textwidth]{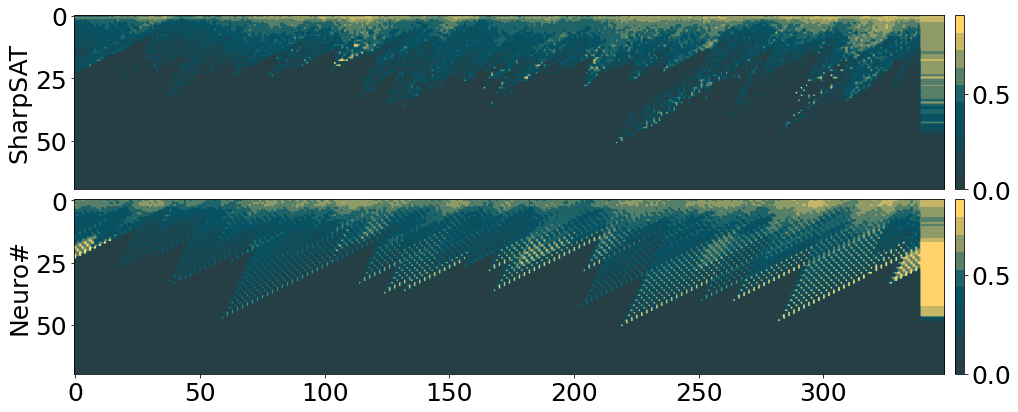} 
\caption{Full-sized variable selection heatmap on dataset \cell[35,348,280]. 
We show the 99th percentile for each row of the heatmap in the last column. 
}
\label{fig:heat35full}
\end{figure}

Formal analysis of the learned policy and its performance improvements is difficult, 
however we can form some high-level conjectures regarding the behaviour of \neurosharp
by how well it decomposes the problem. 
Here we look at \cell. The reason is that this problem has a straightforward encoding that directly relates the CNF representation to an easy-to-visualize evolution grid that coincides with the standard representation of Elementary Cellular Automata. 
The natural way to decompose this problem is to start from the known state $T$ (bottom row) and to ``guess'' the preimage, row by row from the bottom up through variable assignment. Different preimages can be computed \emph{independently} upwards, and indeed, this is how a human would approach the problem. Heat maps in Figure \ref{fig:cells} (c~\&~d) depict the behaviour under \sharpsat{} and \neurosharp respectively. The heat map aligns with the evolution grid, with the terminal state $T$ at the bottom. The hotter coloured cells indicate that, on average, the corresponding variable is branched on earlier by the policy. The cooler colours show that the variable is often selected later or not at all, meaning that its value is often inferred through UP either initially or after some variable assignments. That is why the bottom row $T$ and adjacent rows are completely dark, because they are simplified by the solver before any branching happens. We show the effect of this early simplification on a single formula per dataset in Figure \ref{fig:cells} (b). Notice that in \cell[35 \& 49] the simplification shatters the problem space into few small components (dark triangles), while in \cell[9] which is a more challenging problem, it only chips away a small region of the problem space, leaving it as a single component. Regardless of this, as conjectured, we can see a clear trend with \neurosharp focusing more on branching early on variables of the bottom rows in \cell[9] and in a less pronounced way in \cell[35 \& 49]. Moreover, as more clearly seen in the heatmap for the larger problem in Figure \ref{fig:heat35full}, \neurosharp actually branches early according to the pattern of the rule.

\subsection{Ablation Study}\label{sect:Ablation} 
\begin{figure}[tbh]
    \centering
    \includegraphics[width=0.7\columnwidth]{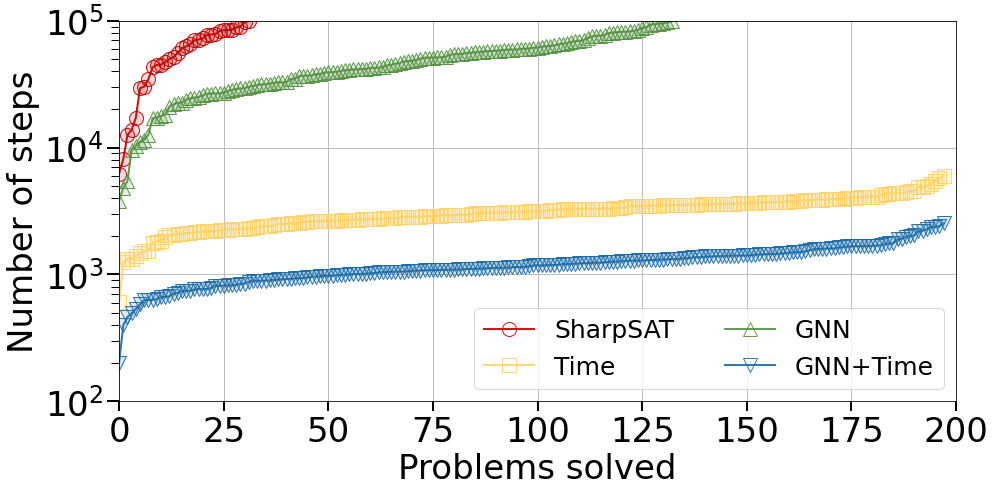}
    \caption{Ablation study on the impact of the ``time'' feature on upward generalization on \grid[10,12].}
    \label{fig:ablation}
\end{figure}

We tested the degree to which the ``time'' feature contributed to the upward generalization performance of \grid{}. We compared three architectures with \sharpsat as the baseline: \begin{enumerate*}[label=\textbf{\arabic*.}]
\item \emph{GNN}: The standard architecture proposed in Section~\ref{subsec:model},
\item \emph{GNN+Time}: Same as \emph{GNN} but with the variable embeddings augmented with the ``time'' feature and \item \emph{Time}: Where no variable embedding is computed and only the ``time'' feature is fed into the policy network
\end{enumerate*}. As depicted in Figure \ref{fig:ablation}, we discovered that ``time'' is responsible for most of the improvement over \sharpsat.  This fact is encouraging, because it demonstrates the potential gains that could be achieved by simply utilizing problem-level data, such as ``time'', that otherwise would have been lost during the CNF encoding. 

\subsection{Discussions}\label{sec:discussion}
We observed varying degrees of success on different problem families. 
This raises the question of what traits make a problem more amenable for improvement via \neurosharp{}. One of the main contributing factors is the model's ability to observe similar components many times during training. In other words, if a problem gets shattered into smaller components either by the initial simplification (e.g., UP) or after a few variable assignments, there is a high chance that the model fits to such distribution of components. If larger problems of the same domain also break down into similar component structures, then \neurosharp can generalize well on them. This explains why sequential problems like \grid{} benefit from our method, as they are naturally decomposable into similar iterations and addition of the ``time'' feature demarcates the boundaries between iterations even more clearer. 

Recently \cite{DBLP:conf/ictai/BliemJ19} showed branching according to the centrality scores of variable nodes of the CNF graph leads to significant performance improvements. We obtained their centrality enhanced version of \sharpsat and compared it with \neurosharp trained on specific problem family. We found that, although centrality enhanced \sharpsat, \neurosharp retained its orders of magnitude superiority over it. This indicates that whatever structure \neurosharp is exploiting from the graph, it is not exclusively centrality. Also we compared the performance of \neurosharp{} against the state-of-the-art \emph{probabilistic} model counter \ganak \cite{DBLP:conf/ijcai/SharmaRSM19}. Note that this solver performs the easier task of providing a model count that is only \emph{probably correct within a given error tolerance}. Thus to make the comparison more fair we set the error tolerance of \ganak to a small value of $10^{-3}$ and observed that its performance was again inferior to \neurosharp{}. An interesting future direction would be to investigate if our method could also be used to customize \ganak's heuristics. Appendix~\ref{appendix:results} includes cactus plots for every problem in our dataset along with comparison to both centrality and \ganak.

\section{Conclusion}\label{sec:conclusion}
We studied the feasibility of enhancing the branching heuristic in propositional model counting via learning. We used solver's branching steps as a measure of its performance and trained our model to minimize that measure. We demonstrated experimentally that the resulting model not only is capable of generalizing to the unseen instances from the same problem distribution, but also maintains its lead relative to \sharpsat on larger problems. For certain problems, this lead widens to a degree that the trained model achieves wall-clock time improvement over the standard heuristic, in spite of the imposed runtime overhead of querying the model. This is an exciting first step and it positions this line of research as a potential path towards building better model counters and thus broadening their application horizon.

\paragraph{Acknowledgements} Dr. Seshia's contribution to this work was funded by DARPA contract FA8750-20-C-0156 (LOGiCS), NSF grant CNS-1545126 (VeHICaL), Microsoft and Intel. Gil Lederman's work was partially funded by NSF award CNS-1836601.

\begin{small}
\bibliography{refs}

\begin{thebibliography}{49}
\providecommand{\natexlab}[1]{#1}
\providecommand{\url}[1]{\texttt{#1}}
\providecommand{\urlprefix}{URL }
\expandafter\ifx\csname urlstyle\endcsname\relax
  \providecommand{\doi}[1]{doi:\discretionary{}{}{}#1}\else
  \providecommand{\doi}{doi:\discretionary{}{}{}\begingroup
  \urlstyle{rm}\Url}\fi

\bibitem[{Abboud, Ceylan, and Lukasiewicz(2020)}]{DBLP:conf/aaai/AbboudCL20}
Abboud, R.; Ceylan, {\.I}.~{\.I}.; and Lukasiewicz, T. 2020.
\newblock Learning to Reason: Leveraging Neural Networks for Approximate {DNF}
  Counting.
\newblock In \emph{The Thirty-Fourth {AAAI} Conference on Artificial
  Intelligence, {AAAI} 2020, The Thirty-Second Innovative Applications of
  Artificial Intelligence Conference}, 3097--3104. {AAAI} Press.
\newblock
  \urlprefix\url{https://aaai.org/ojs/index.php/AAAI/article/view/5705}.

\bibitem[{Amizadeh, Matusevych, and Weimer(2019)}]{amizadeh2018learning}
Amizadeh, S.; Matusevych, S.; and Weimer, M. 2019.
\newblock {Learning To Solve {Circuit-SAT}: An Unsupervised Differentiable
  Approach}.
\newblock In \emph{7th International Conference on Learning Representations,
  {ICLR} 2019, New Orleans, LA, USA, May 6-9, 2019}. OpenReview.net.
\newblock \url{https://openreview.net/forum?id=BJxgz2R9t7}.

\bibitem[{Bacchus, Dalmao, and
  Pitassi(2003{\natexlab{a}})}]{DBLP:conf/focs/BacchusDP03}
Bacchus, F.; Dalmao, S.; and Pitassi, T. 2003{\natexlab{a}}.
\newblock {Algorithms and Complexity Results for {{\#}SAT} and {Bayesian}
  Inference}.
\newblock In \emph{44th Symposium on Foundations of Computer Science {(FOCS}
  2003), 11-14 October 2003, Cambridge, MA, USA, Proceedings}, 340--351. {IEEE}
  Computer Society.
\newblock \doi{10.1109/SFCS.2003.1238208}.
\newblock \url{https://doi.org/10.1109/SFCS.2003.1238208}.

\bibitem[{Bacchus, Dalmao, and
  Pitassi(2003{\natexlab{b}})}]{DBLP:conf/uai/BacchusDP03}
Bacchus, F.; Dalmao, S.; and Pitassi, T. 2003{\natexlab{b}}.
\newblock {Value Elimination: Bayesian Interence via Backtracking Search}.
\newblock In \emph{{UAI} '03, Proceedings of the 19th Conference in Uncertainty
  in Artificial Intelligence, Acapulco, Mexico, August 7-10 2003}. Morgan
  Kaufmann.
\newblock
  \url{https://dslpitt.org/uai/displayArticleDetails.jsp?mmnu=1\&smnu=2\&article\_id=909\&proceeding\_id=19}.

\bibitem[{Bacchus, Dalmao, and Pitassi(2009)}]{DBLP:journals/jair/BacchusDP09}
Bacchus, F.; Dalmao, S.; and Pitassi, T. 2009.
\newblock {Solving {{\#}SAT} and {Bayesian} Inference with Backtracking
  Search}.
\newblock \emph{J. Artif. Intell. Res.} 34.
\newblock \doi{10.1613/jair.2648}.
\newblock \url{https://doi.org/10.1613/jair.2648}.

\bibitem[{Balcan et~al.(2018)Balcan, Dick, Sandholm, and
  Vitercik}]{DBLP:conf/icml/BalcanDSV18}
Balcan, M.; Dick, T.; Sandholm, T.; and Vitercik, E. 2018.
\newblock {Learning to Branch}.
\newblock In \emph{Proceedings of the 35th International Conference on Machine
  Learning, {ICML} 2018, Stockholmsm{\"{a}}ssan, Stockholm, Sweden, July 10-15,
  2018}, volume~80 of \emph{Proceedings of Machine Learning Research}. {PMLR}.
\newblock \url{http://proceedings.mlr.press/v80/balcan18a.html}.

\bibitem[{Bayardo and Pehoushek(2000)}]{DBLP:conf/aaai/Pehoushek00}
Bayardo, Jr., R.~J.; and Pehoushek, J.~D. 2000.
\newblock {Counting Models Using Connected Components}.
\newblock In \emph{Proceedings of the Seventeenth National Conference on
  Artificial Intelligence and Twelfth Conference on on Innovative Applications
  of Artificial Intelligence, July 30 - August 3, 2000, Austin, Texas, {USA}},
  157--162. {AAAI} Press / The {MIT} Press.
\newblock \url{http://www.aaai.org/Library/AAAI/2000/aaai00-024.php}.

\bibitem[{Beyer and Schwefel(2002)}]{DBLP:journals/nc/BeyerS02}
Beyer, H.; and Schwefel, H. 2002.
\newblock {Evolution strategies - A Comprehensive Introduction}.
\newblock \emph{Nat. Comput.} 1(1): 3--52.
\newblock \url{https://doi.org/10.1023/A:1015059928466}.

\bibitem[{Birnbaum and Lozinskii(1999)}]{DBLP:journals/jair/BirnbaumL99}
Birnbaum, E.; and Lozinskii, E.~L. 1999.
\newblock {The Good Old Davis-Putnam Procedure Helps Counting Models}.
\newblock \emph{J. Artif. Intell. Res.} 10: 457--477.
\newblock \url{https://doi.org/10.1613/jair.601}.

\bibitem[{Bliem and J{\"{a}}rvisalo(2019)}]{DBLP:conf/ictai/BliemJ19}
Bliem, B.; and J{\"{a}}rvisalo, M. 2019.
\newblock Centrality Heuristics for Exact Model Counting.
\newblock In \emph{31st {IEEE} International Conference on Tools with
  Artificial Intelligence, {ICTAI} 2019, Portland, OR, USA, November 4-6,
  2019}, 59--63. {IEEE}.
\newblock \urlprefix\url{https://doi.org/10.1109/ICTAI.2019.00017}.

\bibitem[{Chakraborty et~al.(2014)Chakraborty, Fremont, Meel, Seshia, and
  Vardi}]{DBLP:conf/aaai/ChakrabortyFMSV14}
Chakraborty, S.; Fremont, D.~J.; Meel, K.~S.; Seshia, S.~A.; and Vardi, M.~Y.
  2014.
\newblock {Distribution-Aware Sampling and Weighted Model Counting for {SAT}}.
\newblock In \emph{Proceedings of the Twenty-Eighth {AAAI} Conference on
  Artificial Intelligence, July 27 -31, 2014, Qu{\'{e}}bec City, Qu{\'{e}}bec,
  Canada}, 1722--1730. {AAAI} Press.

\bibitem[{Davis, Logemann, and Loveland(1962)}]{DBLP:journals/cacm/DavisLL62}
Davis, M.; Logemann, G.; and Loveland, D.~W. 1962.
\newblock A machine program for theorem-proving.
\newblock \emph{Commun. {ACM}} 5(7): 394--397.
\newblock \doi{10.1145/368273.368557}.
\newblock \urlprefix\url{https://doi.org/10.1145/368273.368557}.

\bibitem[{Domshlak and Hoffmann(2006)}]{DBLP:conf/aips/DomshlakH06}
Domshlak, C.; and Hoffmann, J. 2006.
\newblock Fast Probabilistic Planning through Weighted Model Counting.
\newblock In \emph{Proceedings of the Sixteenth International Conference on
  Automated Planning and Scheduling, {ICAPS} 2006, Cumbria, UK, June 6-10,
  2006}, 243--252. {AAAI}.
\newblock
  \urlprefix\url{http://www.aaai.org/Library/ICAPS/2006/icaps06-025.php}.

\bibitem[{Domshlak and Hoffmann(2007)}]{DBLP:journals/jair/DomshlakH07}
Domshlak, C.; and Hoffmann, J. 2007.
\newblock {Probabilistic Planning via Heuristic Forward Search and Weighted
  Model Counting}.
\newblock \emph{J. Artif. Intell. Res.} 30.
\newblock \doi{10.1613/jair.2289}.
\newblock \url{https://doi.org/10.1613/jair.2289}.

\bibitem[{Gasse et~al.(2019)Gasse, Ch{\'{e}}telat, Ferroni, Charlin, and
  Lodi}]{DBLP:conf/nips/GasseCFC019}
Gasse, M.; Ch{\'{e}}telat, D.; Ferroni, N.; Charlin, L.; and Lodi, A. 2019.
\newblock {Exact Combinatorial Optimization with Graph Convolutional Neural
  Networks}.
\newblock In \emph{Advances in Neural Information Processing Systems 32: Annual
  Conference on Neural Information Processing Systems 2019, NeurIPS 2019, 8-14
  December 2019, Vancouver, BC, Canada}.
\newblock
  \url{http://papers.nips.cc/paper/9690-exact-combinatorial-optimization-with-graph-convolutional-neural-networks}.

\bibitem[{Geffner and Geffner(2018)}]{DBLP:conf/aips/GeffnerG18}
Geffner, T.; and Geffner, H. 2018.
\newblock Compact Policies for Fully Observable Non-Deterministic Planning as
  {SAT}.
\newblock In \emph{Proceedings of the Twenty-Eighth International Conference on
  Automated Planning and Scheduling, {ICAPS} 2018, Delft, The Netherlands, June
  24-29, 2018}, 88--96. {AAAI} Press.

\bibitem[{Gomes, Sabharwal, and Selman(2009)}]{DBLP:series/faia/GomesSS09}
Gomes, C.~P.; Sabharwal, A.; and Selman, B. 2009.
\newblock {Model Counting}.
\newblock In \emph{Handbook of Satisfiability}, volume 185 of \emph{Frontiers
  in Artificial Intelligence and Applications}, 633--654. {IOS} Press.
\newblock \url{https://doi.org/10.3233/978-1-58603-929-5-633}.

\bibitem[{{Gori}, {Monfardini}, and {Scarselli}(2005)}]{gnn_gori}
{Gori}, M.; {Monfardini}, G.; and {Scarselli}, F. 2005.
\newblock {A New Model for Learning in Graph Domains}.
\newblock In \emph{Proceedings. 2005 IEEE International Joint Conference on
  Neural Networks, 2005.}, volume~2.

\bibitem[{Hansknecht, Joormann, and Stiller(2018)}]{hansknecht2018cuts}
Hansknecht, C.; Joormann, I.; and Stiller, S. 2018.
\newblock {Cuts, Primal Heuristics, and Learning to Branch for the
  Time-Dependent Traveling Salesman Problem}.
\newblock Technical report, arXiv.
\newblock \url{https://arxiv.org/abs/1805.01415}.

\bibitem[{Heule, J{\"a}rvisalo, and Suda(2018)}]{heule2018proceedings}
Heule, M.~J.; J{\"a}rvisalo, M.~J.; and Suda, M. 2018.
\newblock Proceedings of SAT Competition 2018: Solver and Benchmark
  Descriptions \url{http://hdl.handle.net/10138/237063}.

\bibitem[{Karp, Luby, and Madras(1989)}]{DBLP:journals/jal/KarpLM89}
Karp, R.~M.; Luby, M.; and Madras, N. 1989.
\newblock Monte-Carlo Approximation Algorithms for Enumeration Problems.
\newblock \emph{J. Algorithms} 10(3): 429--448.
\newblock \doi{10.1016/0196-6774(89)90038-2}.
\newblock \urlprefix\url{https://doi.org/10.1016/0196-6774(89)90038-2}.

\bibitem[{Khalil et~al.(2016)Khalil, Bodic, Song, Nemhauser, and
  Dilkina}]{DBLP:conf/aaai/KhalilBSND16}
Khalil, E.~B.; Bodic, P.~L.; Song, L.; Nemhauser, G.~L.; and Dilkina, B. 2016.
\newblock {Learning to Branch in Mixed Integer Programming}.
\newblock In \emph{Proceedings of the Thirtieth {AAAI} Conference on Artificial
  Intelligence, February 12-17, 2016, Phoenix, Arizona, {USA}}, 724--731.
  {AAAI} Press.
\newblock \url{http://www.aaai.org/ocs/index.php/AAAI/AAAI16/paper/view/12514}.

\bibitem[{Kingma and Ba(2015)}]{kingma2014adam}
Kingma, D.~P.; and Ba, J. 2015.
\newblock {Adam: A Method for Stochastic Optimization}.
\newblock In \emph{3rd International Conference on Learning Representations,
  {ICLR} 2015, San Diego, CA, USA, May 7-9, 2015, Conference Track
  Proceedings}.
\newblock \url{http://arxiv.org/abs/1412.6980}.

\bibitem[{Kurin et~al.(2019)Kurin, Godil, Whiteson, and
  Catanzaro}]{kurin2019improving}
Kurin, V.; Godil, S.; Whiteson, S.; and Catanzaro, B. 2019.
\newblock {Improving {SAT} Solver Heuristics with Graph Networks and
  Reinforcement Learning}.
\newblock \emph{CoRR} abs/1909.11830.
\newblock \url{http://arxiv.org/abs/1909.11830}.

\bibitem[{Lederman et~al.(2020)Lederman, Rabe, Seshia, and Lee}]{LedermanRSL20}
Lederman, G.; Rabe, M.~N.; Seshia, S.; and Lee, E.~A. 2020.
\newblock {Learning Heuristics for Quantified Boolean Formulas through
  Reinforcement Learning}.
\newblock In \emph{8th International Conference on Learning Representations,
  {ICLR} 2020, Addis Ababa, Ethiopia, April 26-30, 2020}. OpenReview.net.
\newblock \url{https://openreview.net/forum?id=BJluxREKDB}.

\bibitem[{Li, Poupart, and van Beek(2011)}]{DBLP:journals/jair/0002PB11}
Li, W.; Poupart, P.; and van Beek, P. 2011.
\newblock {Exploiting Structure in Weighted Model Counting Approaches to
  Probabilistic Inference}.
\newblock \emph{J. Artif. Intell. Res.} 40.
\newblock \url{http://jair.org/papers/paper3232.html}.

\bibitem[{Marques{-}Silva(2018)}]{DBLP:conf/cie/Marques-Silva18}
Marques{-}Silva, J. 2018.
\newblock {Computing with SAT Oracles: Past, Present and Future}.
\newblock In \emph{Sailing Routes in the World of Computation - 14th Conference
  on Computability in Europe, CiE 2018, Kiel, Germany, July 30 - August 3,
  2018, Proceedings}, volume 10936 of \emph{Lecture Notes in Computer Science}.
  Springer.
\newblock \url{https://doi.org/10.1007/978-3-319-94418-0\_27}.

\bibitem[{Meel and Akshay(2020)}]{DBLP:journals/corr/abs-2004-14692}
Meel, K.~S.; and Akshay, S. 2020.
\newblock {Sparse Hashing for Scalable Approximate Model Counting: Theory and
  Practice}.
\newblock \emph{CoRR} abs/2004.14692.
\newblock \url{https://arxiv.org/abs/2004.14692}.

\bibitem[{Oztok and Darwiche(2015)}]{DBLP:conf/ijcai/OztokD15}
Oztok, U.; and Darwiche, A. 2015.
\newblock {A Top-Down Compiler for Sentential Decision Diagrams}.
\newblock In \emph{Proceedings of the Twenty-Fourth International Joint
  Conference on Artificial Intelligence, {IJCAI} 2015, Buenos Aires, Argentina,
  July 25-31, 2015}, 3141--3148. {AAAI} Press.
\newblock \url{http://ijcai.org/Abstract/15/443}.

\bibitem[{Robertson and Seymour(1991)}]{DBLP:journals/jct/RobertsonS91}
Robertson, N.; and Seymour, P.~D. 1991.
\newblock {Graph Minors. X. Obstructions to Tree-Decomposition}.
\newblock \emph{J. Comb. Theory, Ser. {B}} 52(2): 153--190.
\newblock \url{https://doi.org/10.1016/0095-8956(91)90061-N}.

\bibitem[{Robertson and Seymour(2010)}]{DBLP:journals/jct/RobertsonS10}
Robertson, N.; and Seymour, P.~D. 2010.
\newblock {Graph Minors {XXIII.} Nash-Williams' Immersion Conjecture}.
\newblock \emph{J. Comb. Theory, Ser. {B}} 100(2): 181--205.
\newblock \url{https://doi.org/10.1016/j.jctb.2009.07.003}.

\bibitem[{Salimans et~al.(2017)Salimans, Ho, Chen, and
  Sutskever}]{DBLP:journals/corr/SalimansHCS17}
Salimans, T.; Ho, J.; Chen, X.; and Sutskever, I. 2017.
\newblock {Evolution Strategies as a Scalable Alternative to Reinforcement
  Learning}.
\newblock \emph{CoRR} abs/1703.03864.
\newblock \urlprefix\url{http://arxiv.org/abs/1703.03864}.

\bibitem[{Sang et~al.(2004)Sang, Bacchus, Beame, Kautz, and
  Pitassi}]{DBLP:conf/sat/SangBBKP04}
Sang, T.; Bacchus, F.; Beame, P.; Kautz, H.~A.; and Pitassi, T. 2004.
\newblock {Combining Component Caching and Clause Learning for Effective Model
  Counting}.
\newblock In \emph{{SAT} 2004 - The Seventh International Conference on Theory
  and Applications of Satisfiability Testing, 10-13 May 2004, Vancouver, BC,
  Canada, Online Proceedings}.
\newblock \url{http://www.satisfiability.org/SAT04/programme/21.pdf}.

\bibitem[{Sang, Beame, and Kautz(2005{\natexlab{a}})}]{DBLP:conf/sat/SangBK05}
Sang, T.; Beame, P.; and Kautz, H.~A. 2005{\natexlab{a}}.
\newblock {Heuristics for Fast Exact Model Counting}.
\newblock In \emph{Theory and Applications of Satisfiability Testing, 8th
  International Conference, {SAT} 2005, St. Andrews, UK, June 19-23, 2005,
  Proceedings}, volume 3569 of \emph{Lecture Notes in Computer Science},
  226--240. Springer.
\newblock \url{https://doi.org/10.1007/11499107\_17}.

\bibitem[{Sang, Beame, and Kautz(2005{\natexlab{b}})}]{DBLP:conf/aaai/SangBK05}
Sang, T.; Beame, P.; and Kautz, H.~A. 2005{\natexlab{b}}.
\newblock {Performing Bayesian Inference by Weighted Model Counting}.
\newblock In \emph{Proceedings, The Twentieth National Conference on Artificial
  Intelligence and the Seventeenth Innovative Applications of Artificial
  Intelligence Conference, July 9-13, 2005, Pittsburgh, Pennsylvania, {USA}}.
  {AAAI} Press / The {MIT} Press.
\newblock \url{http://www.aaai.org/Library/AAAI/2005/aaai05-075.php}.

\bibitem[{Scarselli et~al.(2009)Scarselli, Gori, Tsoi, Hagenbuchner, and
  Monfardini}]{ScarselliGTHM09}
Scarselli, F.; Gori, M.; Tsoi, A.~C.; Hagenbuchner, M.; and Monfardini, G.
  2009.
\newblock {The Graph Neural Network Model}.
\newblock \emph{{IEEE} Trans. Neural Networks} 20(1): 61--80.
\newblock \url{https://doi.org/10.1109/TNN.2008.2005605}.

\bibitem[{Selman, Kautz, and Cohen(1993)}]{selman1993local}
Selman, B.; Kautz, H.~A.; and Cohen, B. 1993.
\newblock {Local Search Strategies for Satisfiability Testing}.
\newblock In \emph{Cliques, Coloring, and Satisfiability, Proceedings of a
  {DIMACS} Workshop, New Brunswick, New Jersey, USA, October 11-13, 1993},
  volume~26 of \emph{{DIMACS} Series in Discrete Mathematics and Theoretical
  Computer Science}, 521--531. {DIMACS/AMS}.
\newblock \url{https://doi.org/10.1090/dimacs/026/25}.

\bibitem[{Selsam and Bj{\o}rner(2019)}]{selsam2019neurocore}
Selsam, D.; and Bj{\o}rner, N. 2019.
\newblock {{NeuroCore}: Guiding High-Performance {SAT} Solvers with Unsat-Core
  Predictions}.
\newblock \emph{CoRR} abs/1903.04671.
\newblock \url{http://arxiv.org/abs/1903.04671}.

\bibitem[{Selsam et~al.(2019)Selsam, Lamm, B{\"{u}}nz, Liang, de~Moura, and
  Dill}]{selsam2018learning}
Selsam, D.; Lamm, M.; B{\"{u}}nz, B.; Liang, P.; de~Moura, L.; and Dill, D.~L.
  2019.
\newblock {Learning a {SAT} Solver from Single-Bit Supervision}.
\newblock In \emph{7th International Conference on Learning Representations,
  {ICLR} 2019, New Orleans, LA, USA, May 6-9, 2019}. OpenReview.net.
\newblock \url{https://openreview.net/forum?id=HJMC\_iA5tm}.

\bibitem[{Sharma et~al.(2019)Sharma, Roy, Soos, and
  Meel}]{DBLP:conf/ijcai/SharmaRSM19}
Sharma, S.; Roy, S.; Soos, M.; and Meel, K.~S. 2019.
\newblock {GANAK:} {A} Scalable Probabilistic Exact Model Counter.
\newblock In \emph{Proceedings of the Twenty-Eighth International Joint
  Conference on Artificial Intelligence, {IJCAI} 2019, Macao, China, August
  10-16, 2019}, 1169--1176. ijcai.org.
\newblock \urlprefix\url{https://doi.org/10.24963/ijcai.2019/163}.

\bibitem[{Thurley(2006)}]{thurley2006sharpsat}
Thurley, M. 2006.
\newblock {SharpSAT - Counting Models with Advanced Component Caching and
  Implicit BCP}.
\newblock In Biere, A.; and Gomes, C.~P., eds., \emph{Theory and Applications
  of Satisfiability Testing - {SAT} 2006, 9th International Conference,
  Seattle, WA, USA, August 12-15, 2006, Proceedings}, volume 4121 of
  \emph{Lecture Notes in Computer Science}, 424--429. Springer.
\newblock \url{https://doi.org/10.1007/11814948\_38}.

\bibitem[{Toda(1991)}]{DBLP:journals/siamcomp/Toda91}
Toda, S. 1991.
\newblock {PP is as Hard as the Polynomial-Time Hierarchy}.
\newblock \emph{{SIAM} J. Comput.} 20(5): 865--877.
\newblock \url{https://doi.org/10.1137/0220053}.

\bibitem[{Vazquez{-}Chanlatte et~al.(2018)Vazquez{-}Chanlatte, Jha, Tiwari, Ho,
  and Seshia}]{vazquez-neurips18}
Vazquez{-}Chanlatte, M.; Jha, S.; Tiwari, A.; Ho, M.~K.; and Seshia, S.~A.
  2018.
\newblock {Learning Task Specifications from Demonstrations}.
\newblock In \emph{Advances in Neural Information Processing Systems 31: Annual
  Conference on Neural Information Processing Systems 2018, NeurIPS 2018, 3-8
  December 2018, Montr{\'{e}}al, Canada}, 5372--5382.
\newblock
  \url{http://papers.nips.cc/paper/7782-learning-task-specifications-from-demonstrations}.

\bibitem[{Vazquez{-}Chanlatte, Rabe, and Seshia(2019)}]{gridencoding}
Vazquez{-}Chanlatte, M.; Rabe, M.~N.; and Seshia, S.~A. 2019.
\newblock {A Model Counter's Guide to Probabilistic Systems}.
\newblock \emph{CoRR} abs/1903.09354.
\newblock \url{http://arxiv.org/abs/1903.09354}.

\bibitem[{Vemula, Sun, and Bagnell(2019)}]{vemula2019contrasting}
Vemula, A.; Sun, W.; and Bagnell, J.~A. 2019.
\newblock {Contrasting Exploration in Parameter and Action Space: {A}
  Zeroth-Order Optimization Perspective}.
\newblock In \emph{The 22nd International Conference on Artificial Intelligence
  and Statistics, {AISTATS} 2019, 16-18 April 2019, Naha, Okinawa, Japan},
  volume~89 of \emph{Proceedings of Machine Learning Research}, 2926--2935.
  {PMLR}.
\newblock \url{http://proceedings.mlr.press/v89/vemula19a.html}.

\bibitem[{Wierstra et~al.(2014)Wierstra, Schaul, Glasmachers, Sun, Peters, and
  Schmidhuber}]{DBLP:journals/jmlr/WierstraSGSPS14}
Wierstra, D.; Schaul, T.; Glasmachers, T.; Sun, Y.; Peters, J.; and
  Schmidhuber, J. 2014.
\newblock {Natural Evolution Strategies}.
\newblock \emph{J. Mach. Learn. Res.} 15(1): 949--980.
\newblock \url{http://dl.acm.org/citation.cfm?id=2638566}.

\bibitem[{Xu et~al.(2019)Xu, Hu, Leskovec, and Jegelka}]{xu2018powerful}
Xu, K.; Hu, W.; Leskovec, J.; and Jegelka, S. 2019.
\newblock {How Powerful are Graph Neural Networks?}
\newblock In \emph{7th International Conference on Learning Representations,
  {ICLR} 2019, New Orleans, LA, USA, May 6-9, 2019}. OpenReview.net.
\newblock \url{https://openreview.net/forum?id=ryGs6iA5Km}.

\bibitem[{Xu et~al.(2008)Xu, Hutter, Hoos, and Leyton{-}Brown}]{xu2008satzilla}
Xu, L.; Hutter, F.; Hoos, H.~H.; and Leyton{-}Brown, K. 2008.
\newblock {SATzilla: Portfolio-based Algorithm Selection for {SAT}}.
\newblock \emph{J. Artif. Intell. Res.} 32: 565--606.
\newblock \url{https://doi.org/10.1613/jair.2490}.

\bibitem[{Yolcu and P{\'{o}}czos(2019)}]{YolcuP19}
Yolcu, E.; and P{\'{o}}czos, B. 2019.
\newblock {Learning Local Search Heuristics for Boolean Satisfiability}.
\newblock In \emph{Advances in Neural Information Processing Systems 32: Annual
  Conference on Neural Information Processing Systems 2019, NeurIPS 2019, 8-14
  December 2019, Vancouver, BC, Canada}, 7990--8001.
\newblock
  \url{http://papers.nips.cc/paper/9012-learning-local-search-heuristics-for-boolean-satisfiability}.

\end{thebibliography}
\end{small}


\appendix
\appendix
\section{\#SAT Algorithms} \label{appendix:algorithms}
In this section we provide some more details about exact algorithms
for solving \#SAT, see \cite{DBLP:journals/jair/BacchusDP09} for the
full formal details including all proofs.

The simplest algorithm for \#SAT is to extend the backtracking search
DPLL algorithm to make it explore the full set of truth
assignments. This is the basis of the CDP solver presented in
\cite{DBLP:journals/jair/BirnbaumL99}, shown in
Algorithm~\ref{alg:cdp}. In particular, when the current formula
contains an empty clause it has zero models, and when it contains no
clauses each of the remaining $k$ unset variables can be assigned \textbf{true} or
\textbf{false} so there are $2^k$
models (line \ref{alg:cdp:no_clause}).

This algorithm is not very efficient, running in time $2^{\Theta(n)}$
where $n$ is the number of variables in the input formula. Note that
the algorithm is actually a class of algorithms each determined by the
procedure used to select the next literal to branch on. The complexity
bound is strong in the sense that no matter how the branching
decisions are made, we can find a sequence of input formulas on which
the algorithm will take time exponential in $n$ as the formulas get
larger.

Breaking the formula into components and solving each component
separately is an approach suggested by
\citet{DBLP:conf/aaai/Pehoushek00} and used in the \textproc{Relsat} solver. This
approach is shown in Algorithm~\ref{alg:relsat}. This algorithm works
on one component at a time and is identical to \textproc{\#DPLLCache}
(Algorithm~\ref{alg:sharp_dpll}) except that caching is not used.

Breaking the formula into components can yield considerable speedups
depending on $n_0$, the number of variables needed to be set before the formula is
broken into components. If we consider a hypergraph in which every
variable is a node and every clause is a hyperedge over the variables
mentioned in the clause, then the branch-width
\cite{DBLP:journals/jct/RobertsonS91} of this hypergraph provides an
upper bound on $n_0$.
As a result we can
obtain a better upper bound on the run time of \textproc{Relsat} of $n^{O(w)}$
where $w$ is the branch-width of the input's hypergraph. However, this
run time will only be achieved if the branching decisions are made in
an order that respects the branch decomposition with width $w$. In
particular, there exists a sequence of branching decisions achieving a run time
of $n^{O(w)}$. Computing that sequence would
require time $n^{O(1)}2^{O(w)}$ \cite{DBLP:journals/jct/RobertsonS10}, hence a run time of $n^{O(w)}$ can be achieved.

Finally, if component caching is used we obtain
Algorithm~\ref{alg:sharp_dpll} which has a better upper bound of
$2^{O(w)}$.  Again this run time can be achieved with a $n^{O(1)}2^{O(w)}$ computation of an appropriate sequence of branching decisions.

In practice, the branch-width of most instances is very large, making a
run time of $2^{O(w)}$ infeasible. Computing a branching sequence to achieve that run time is also infeasible. Fortunately, in practical instances
unit propagation is also very powerful. This means that making only a
few decisions ($\mbox{} < w$) often allows unit propagation to set $w$
or more variables thus breaking the formula apart into separate
components. Furthermore, most instances are falsified by a large
proportion of their truth assignments. This makes clause learning an
effective addition to \#SAT solvers, as with it the solver can more
effectively traverse the non-solution space.

In sum, for \#SAT solvers the branching decisions try to achieve
complex and sometimes contradictory objectives. Making decisions that
split the formula into larger components near the top of the search
tree (i.e., after only a few decisions are made) allows greater
speedups, while generating many small components near the bottom of
the search trees (i.e., after many decision are made) does not help
the solver. Making decisions that generate the same components under
different branches allows more effective use of the cache. And making
decisions that allow the solver to learn more effective clauses allows
the solver to more efficiently traverse the often large space of
non-solutions. 

\begin{figure}[tb]
\begin{minipage}{0.45\textwidth}
\algrenewcommand\algorithmicindent{1.0em}
\begin{algorithm}[H]
\begin{algorithmic}[1]
\Function{CDP}{$\phi$}
  \If {$\phi$ contains an empty clause}
     \State \Return 0
  \EndIf
  \If {$\phi$ contains no clauses}
     \State $k$ = \# of unset variables
     \State \Return $2^{k}$ \label{alg:cdp:no_clause}
  \EndIf
  \State Pick a literal $l \in \phi$
  \State \Return \textproc{CDP}(\textproc{UP}($\phi$, $l$)) +  \textproc{CDP}(\textproc{UP}($\phi$, $\lnot l$))
\EndFunction
\end{algorithmic}
\caption{DPLL extended to count all
  solutions (CDP)}\label{alg:cdp}
\end{algorithm}
\vspace{-1em}
\begin{algorithm}[H]
\begin{algorithmic}[1]
\Function{Relsat}{$\phi$}
  \State Pick a literal $l \in \phi$ 
  \State \#$l$ = \textproc{CountSide}($\phi$, $l$)
  \State \#$\lnot l$ = \textproc{CountSide}($\phi$, $\lnot l$)
  \State \Return \#$l$ + \#$\lnot l$
\EndFunction
\Statex
\Function{CountSide}{$\phi$, $l$}
  \State $\phi_l = \mbox{}$ \textrm{UP}($\phi$, $l$)
  \If {$\phi_l$ contains an empty clause}
     \State \Return 0
  \EndIf
  \If {$\phi_l$ contains no clauses}
     \State $k$ = \# of unset variables
     \State \Return $2^{k}$
  \EndIf
  \State $K = \textproc{findComponents}(\phi_l)$ 
  \State \Return $\prod_{\kappa\in K} \textproc{Relsat}(\kappa)$
\EndFunction
\end{algorithmic}
\caption{Using Components}\label{alg:relsat}
\end{algorithm}
\end{minipage}
\end{figure}

\section{More on Datasets}\label{appendix:dataset}

\begin{figure*}[tbh]
\centering
\begin{tabular}{cccc}
\includegraphics[width=0.2\textwidth]{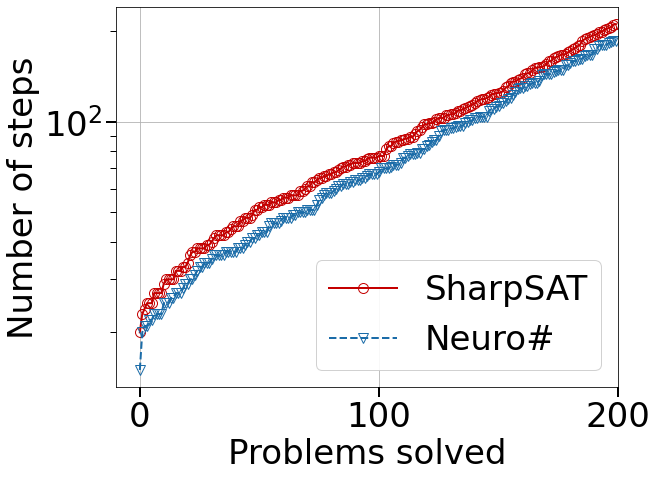}&
\includegraphics[width=0.2\textwidth]{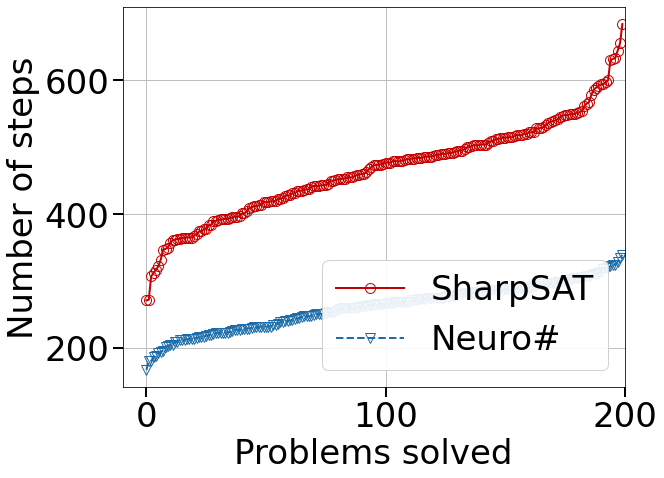}&
\includegraphics[width=0.2\textwidth]{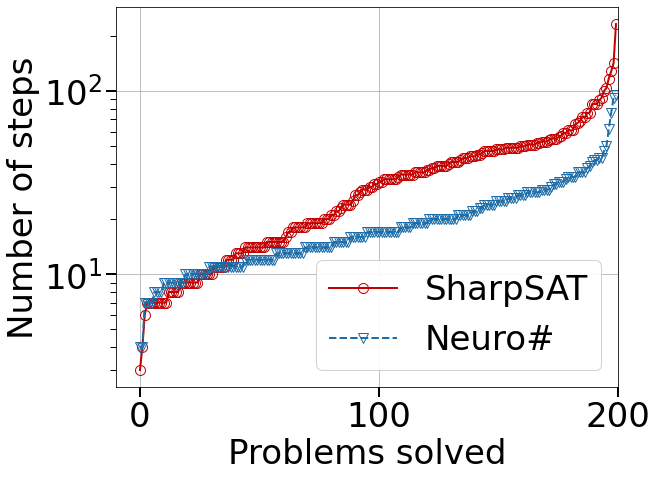}&
\includegraphics[width=0.2\textwidth]{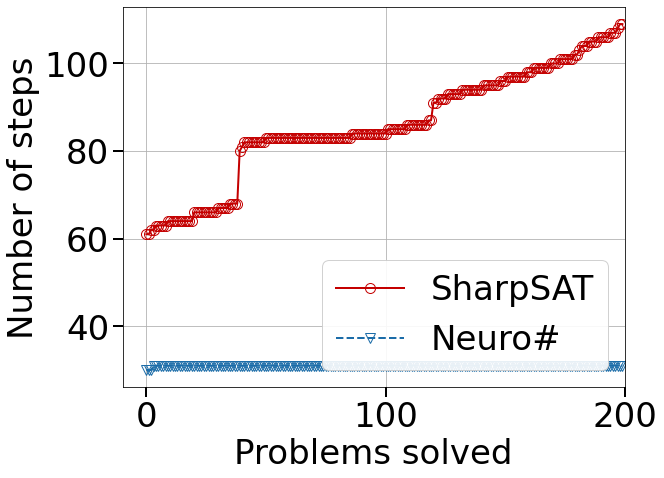}\\
(a) \sudoku{} & (b) \queens & (c) \sha & (d) \island \\[0.3cm]

\multicolumn{4}{c}{
\begin{tabular}{ccc}
\includegraphics[width=0.2\textwidth]{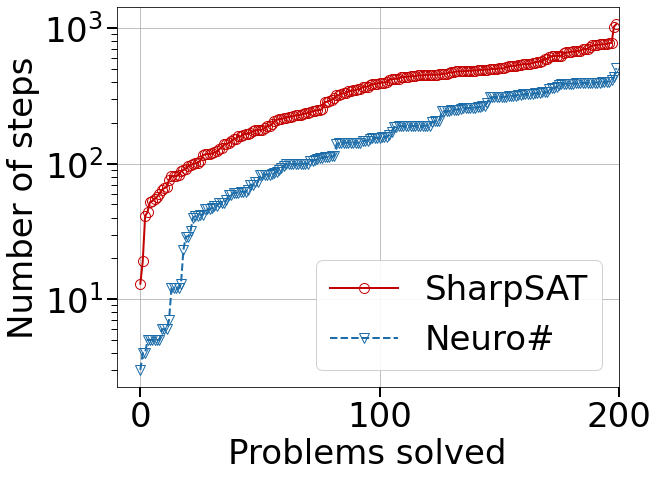}&
\includegraphics[width=0.2\textwidth]{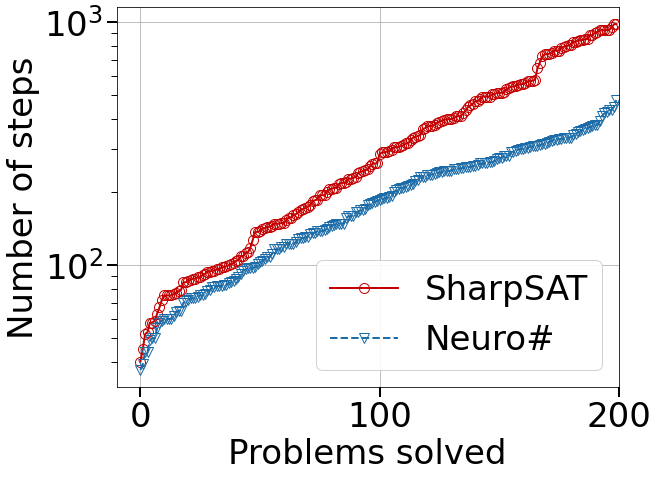}&
\includegraphics[width=0.2\textwidth]{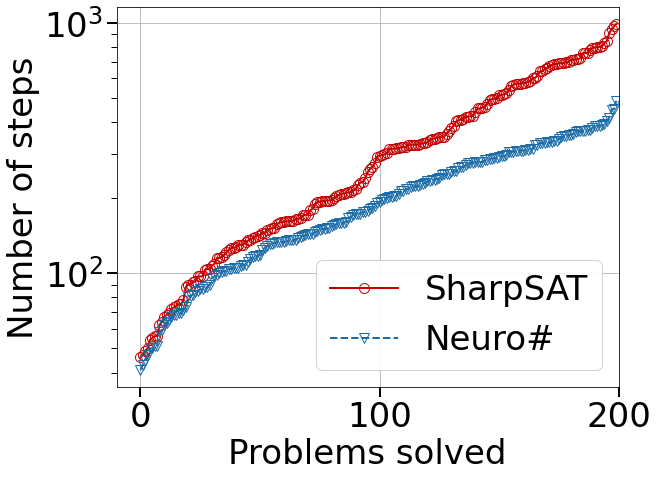}\\
(e) \cell[9] & (f) \cell[35] & (g) \cell[49]\\[0.3cm]
\end{tabular}
}\\

\multicolumn{4}{c}{
\begin{tabular}{ccc}
\includegraphics[width=0.2\textwidth]{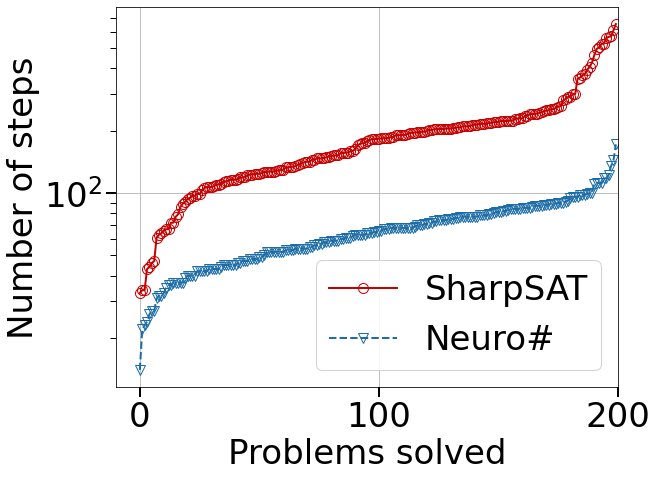}&
\includegraphics[width=0.2\textwidth]{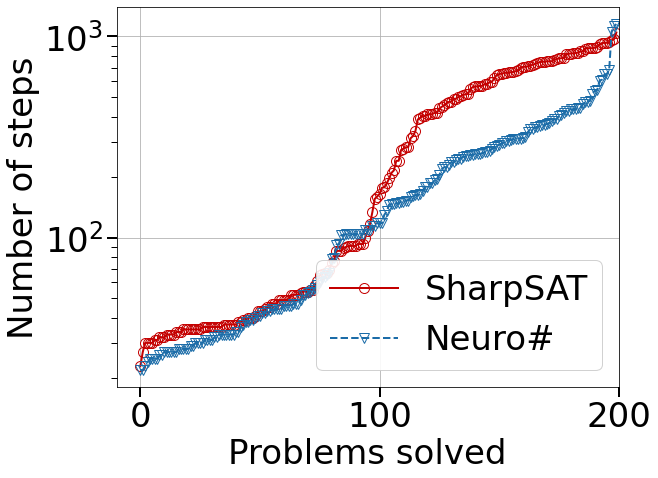}&
\includegraphics[width=0.2\textwidth]{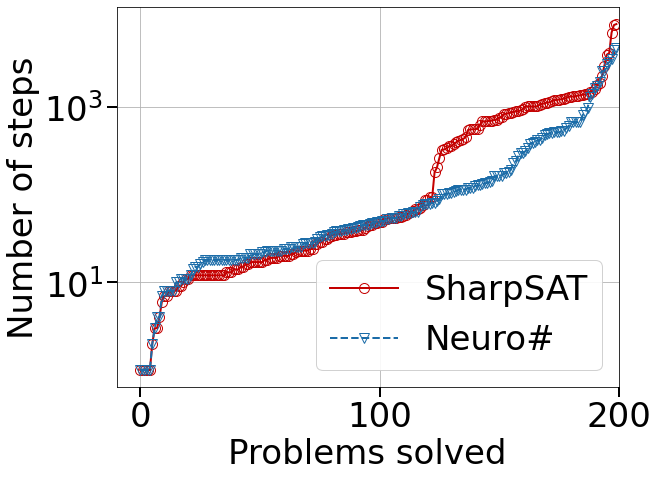}\\
(h) \grid{} & (i) \arith{} & (j) \itarith{}\\
\end{tabular}
}\\
\end{tabular}

\caption{Cactus Plot -- \neurosharp outperforms \sharpsat{} on all i.i.d benchmarks (lower and to the right is better). A cut-off of 100k steps was imposed though both solvers managed to solve the datasets in less than that many steps.}
\label{fig:cactus_same_size}
\end{figure*}

\begin{figure*}[tb]
\centering
\begin{tabular}{c|ccc}
\includegraphics[width=2.95cm]{figures/cactus_cell49_128_110.png}&
\includegraphics[width=2.95cm]{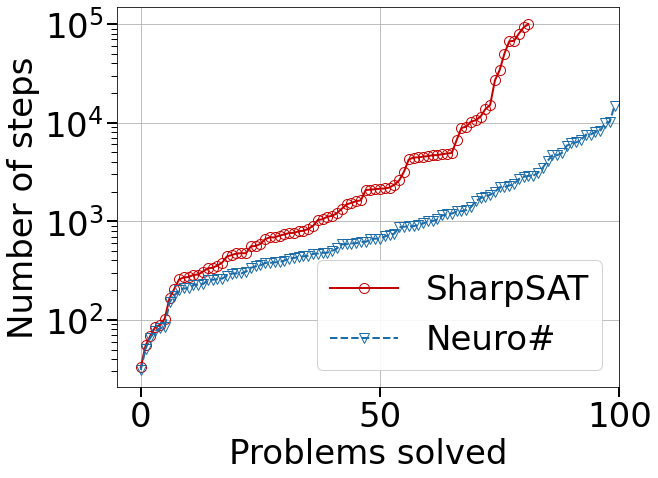}&
\includegraphics[width=2.95cm]{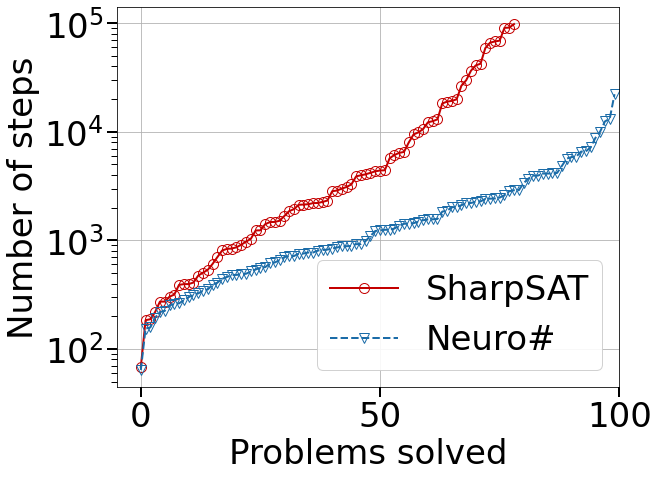}&
\includegraphics[width=2.95cm]{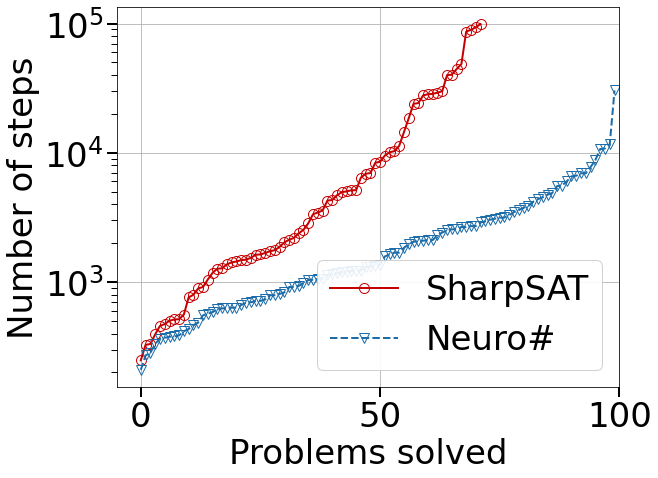}\\
\cell[49,128,110]&\cell[49,192,128]&\cell[49,256,200]&\cell[49,348,280]\\\\


\includegraphics[width=2.95cm]{figures/cactus_grid_10_5.png}&
\includegraphics[width=2.95cm]{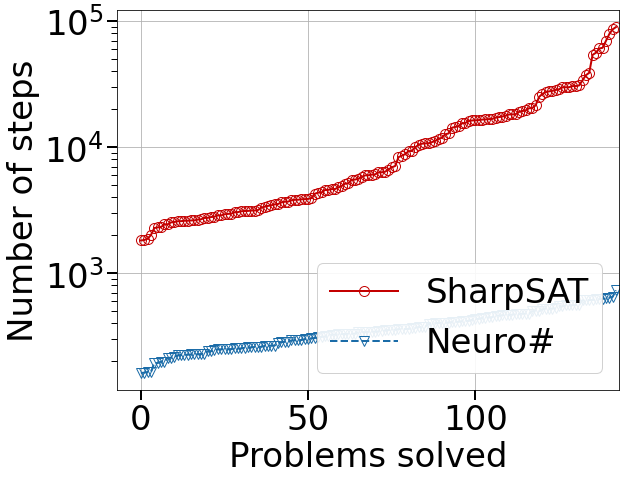}&
\includegraphics[width=2.95cm]{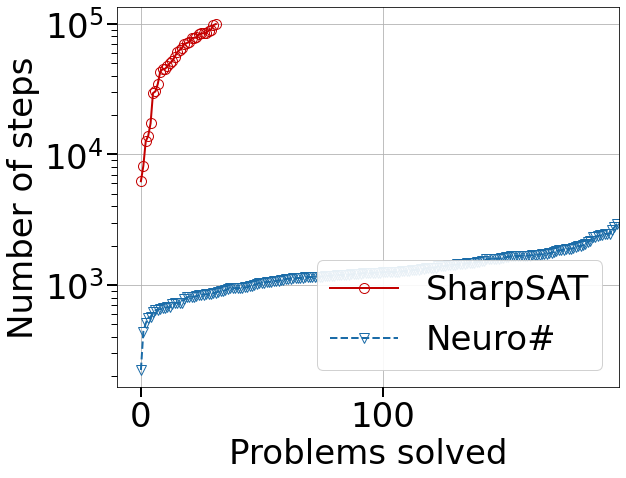}&
\includegraphics[width=2.95cm]{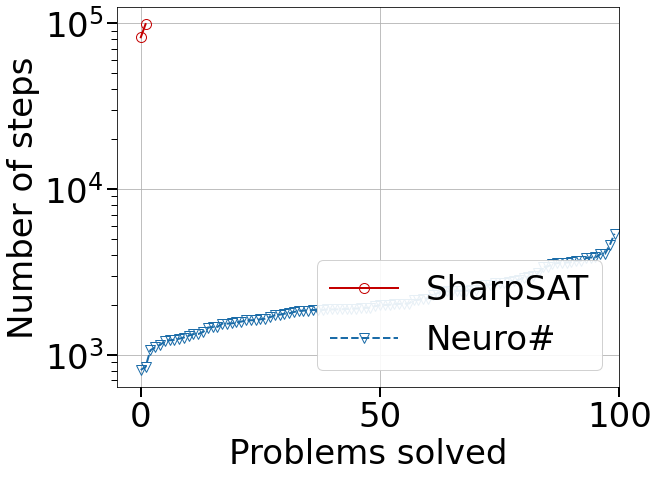}\\
\small{\grid[10,5]}&\small{\grid[10,10]}&\small{\grid[10,12]}&\small{\grid[10,14]}\\
(a)&&(b)&\\
\end{tabular}

\caption{Cactus Plot: \neurosharp maintains its lead over \sharpsat{} on larger datasets (lower and to the right is better). A cut-off of 100k steps was imposed. \textbf{(a)} i.i.d. generalization; \textbf{(b)} Upward generalization of the model trained on \cell[49,128,110] (top row) and \grid[10,5] (bottom row) over larger datasets.}
\label{fig:cactus_grid}
\end{figure*}

Here we provide more detailed information about the problems used in the main document:
\begin{itemize}
\item \sudoku[n,k]: Randomly generated partially filled $n\times n$ Sudoku problems ($n\in\{9, 16\}$) with multiple solutions, where $k$ is the number of revealed squares. 

Sudoku problems are typically designed to have only one solution but as our goal is to improve a model counter, we relaxed this requirement to count the number of solutions instead.

\item \queens[n,k]: Blocked N-Queens problem of size $n$ with $k$ randomly blocked squares on the chess board. This is a standard SAT problem and the task is to count the number of ways to place the $n$ queens on the remaining $n^2-k$ squares such that no two queens attack each other.

\item \sha[n,k]: SHA-1 preimage attack of randomly generated messages of size $n$ with $k$ randomly chosen bits fixed. This problem was taken from SATRACE~2019 
and we used the \textsc{CGen}\footnote{CGEN: \texttt{https://github.com/vsklad/cgen}} tool to generate our instances.

\item \island[n, m]: This dataset was introduced by \cite{DBLP:conf/aips/GeffnerG18} as a \emph{Fully-Observable Non-Deterministic} (FOND) planning problem. There are two grid-like islands of size $n \times n$, each connected by a bridge. The agent is placed at a random location in island 1 and the goal is for it to go to another randomly selected location in island 2. The short (non-deterministic) way is to swim from island 1 to 2, where the agent may drown, and the long way is to go to the bridge and cross it. Crossing the bridge is only possible if no animals are blocking it, otherwise the agent has to move the animals away from the bridge before it can cross it. The $m$ animals are again randomly positioned on the two grids. We used the generative process of \cite{DBLP:conf/aips/GeffnerG18} to encode compact policies for this task in SAT.

\item \arith[n,d,w]: Randomly generated arithmetic sentences of the form $e_1\prec e_2$, where $\prec\in \{\leq, \geq, <, >, =, \neq \}$ and $e_1, e_2$ are expressions of maximum depth $d$ over $n$ binary vector variables of size $w$, random constants and operators ($+, -, \land, \lor, \neg, \mathrm{XOR}, |\cdot| $). The problem is to count the number of integer solutions to the resulting relation in $([0,2^w]\cap \mathbb{Z})^n$.

\item \itarith[s,i]: Randomly generated arithmetic expression circuits of size $s$ (word size fixed at 8). Effectively implementing a function with input and output of a single word. This function is composed $i$ times. We choose a random word $c$, and count the number of inputs such that the output is less than $c$. Formally, if the random circuit is denoted by $f$, we compute $\vert\{x\vert f^i(x)<c\}\vert$.
\end{itemize}

\section{More on the Results}\label{appendix:results}

In this section we present a more elaborate discussion of our results. 
Aggregated measures of performance, such as average number of decisions (Table \ref{table:res} \& \ref{table:res-upwards}) only give us an overall indication of \neurosharp's lead compared to \sharpsat{} and as such, they are incapable of showing whether it is performing better on easier or harder instances in the dataset. Cactus plots are the standard way of comparing solver performances in the SAT community. Although typically used to compare the wall-clock time (bottom row of Figure \ref{fig:wall_clock}), here we use them to compare the number of steps (i.e., branching decisions). 

\paragraph{Hardware Infrastructure}
We used a small cluster of 3 nodes each with an AMD Ryzen Threadripper 2990WX processor with 32 cores (64 threads) and 128GB of memory. We trained for an average of 10 hours on a dataset of 1000 instances for each problem. For testing the wall-clock time we ran the problems sequentially, to avoid any random interference due to parallelism.

\paragraph{Range of Hyperparameters} The model is relatively ``easy'' to train. The criteria for choosing the hyperparameters was the performance of generalization on i.i.d test set, i.e, lowest possible average number of branching decisions. Once calibrated on the first dataset (\cell[35]), we were able to train all models on all datasets without further hyperparameters tuning. The minimal number of episodes per optimization step that worked was 12.
We tested a few different GNN architectures, and none was clearly superior over the others. We also varied the number of GNN message-passing iterations but going beyond 2 iterations had a negative effect (on i.i.d generalization) so we settled on 2.

\subsection{I.I.D. Generalization}\label{appendix:iid}
Figure \ref{fig:cactus_same_size} shows cactus plots for all of the i.i.d. benchmark problems. Unsurprisingly, the improvements on \sudoku{} are relatively modest, albeit consistent across the dataset. On all \cell{} datasets, and \grid{}, a superlinear growth is observed with \neurosharp's lead over \sharpsat{} growing as the problems get more difficult (moving right along the $x$ axis). The problems of the \island i.i.d dataset all had the same model counts and they were isomorphic to one another. Because \neurosharp operates on the graph of the problem, it was capable of utilizing this fact and solve all problems in the same number of steps. However we observe that \sharpsat{}'s performance is function of variable ids and clauses orderings of the input CNF, and thus isomorphic problems are solved in different number of steps. Lastly, on \arith{}, \neurosharp does better almost universally, except near the 100 problems mark and at the very end.

\subsection{Upwards Generalization}\label{appendix:upwards}
On some datasets, namely \cell[49] and \grid{}, \neurosharp's lead over \sharpsat{} becomes more pronounced as we test the upwards generalization (using the model trained on smaller instances and testing on larger ones).
Cactus plots of Figure \ref{fig:cactus_grid} show this effect clearly for these datasets. In each figure, the i.i.d. plot is included as a reference on the left and on the right the plots for test sets with progressively larger instances are depicted. 

\begin{figure*}[t]
\centering
\begin{tabular}{ccc}
\includegraphics[width=0.30\textwidth]{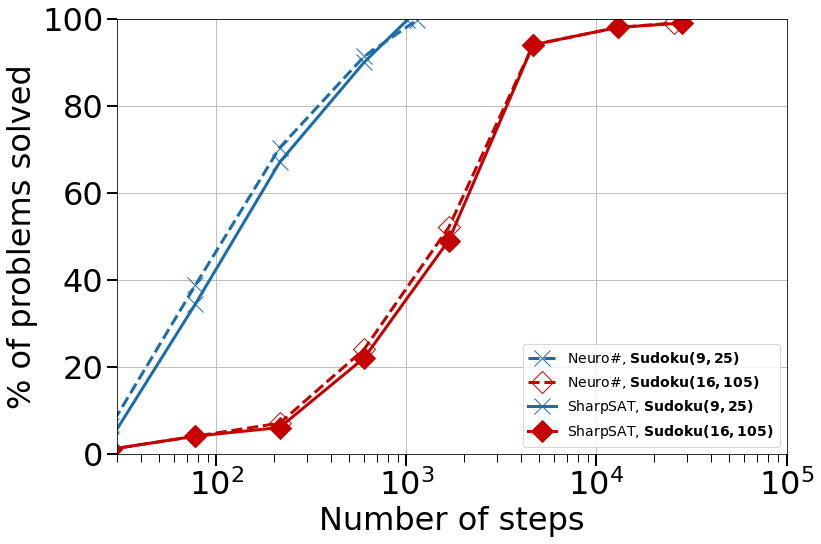}&
\includegraphics[width=0.30\textwidth]{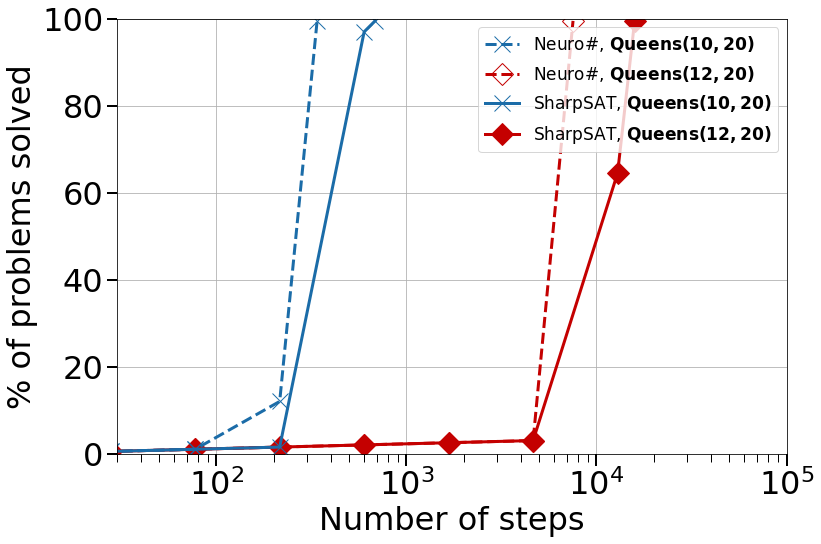}&
\includegraphics[width=0.30\textwidth]{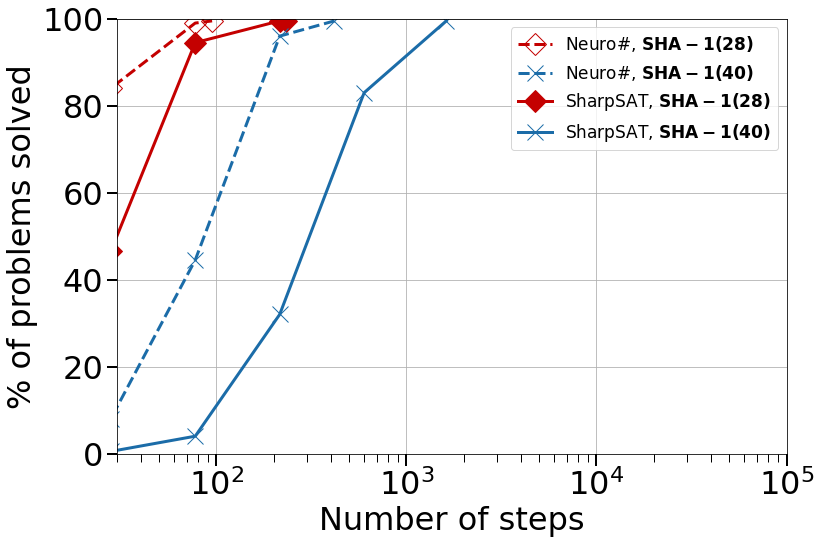}\\
(a) \sudoku{} & (b) \queens[] & (c) \sha\\[0.4cm]

\includegraphics[width=0.30\textwidth]{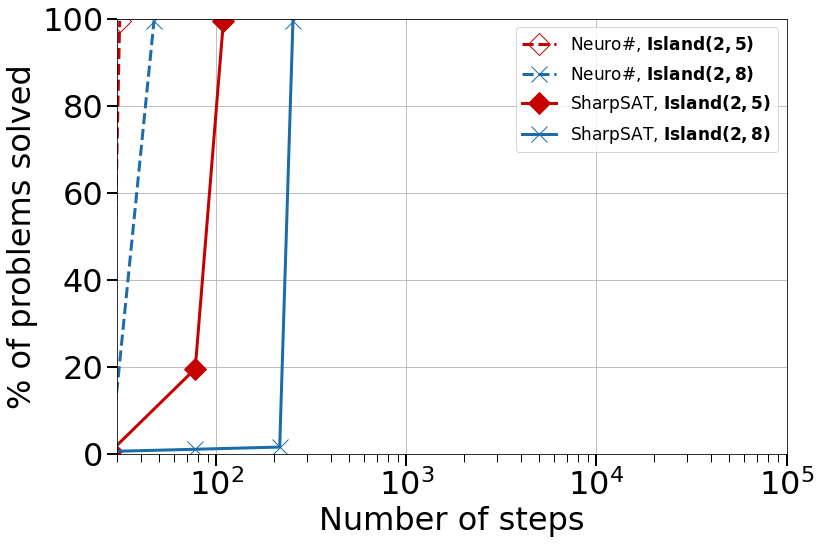}&
\includegraphics[width=0.30\textwidth]{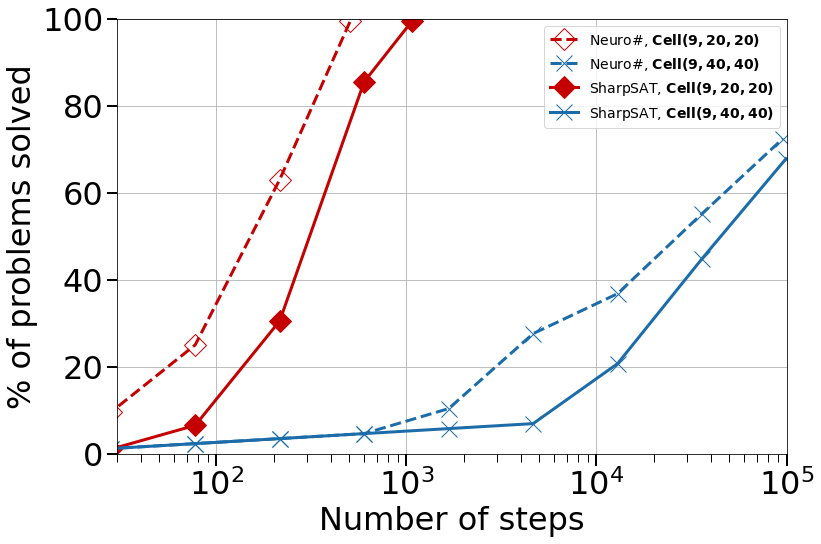}&
\includegraphics[width=0.30\textwidth]{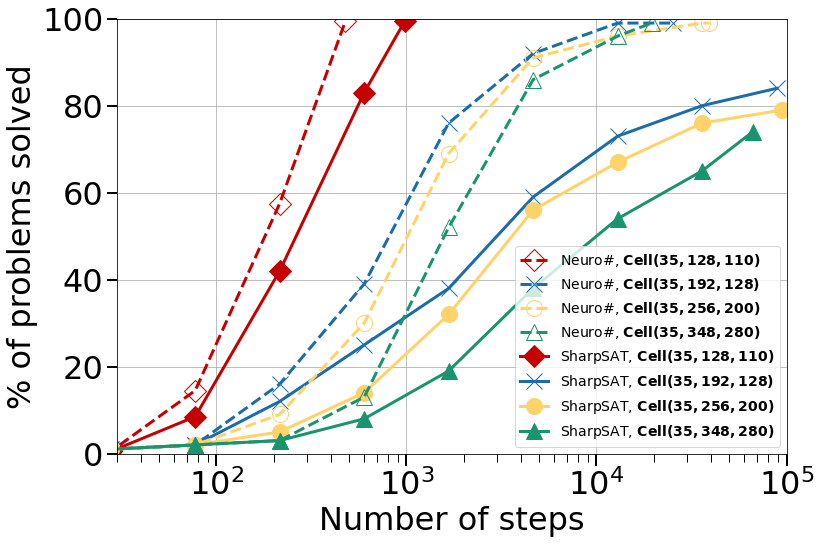}\\
(d) \island & (e) \cell[9] & (f) cell[35]\\[0.4cm]

\includegraphics[width=0.30\textwidth]{figures/percent_cell_49.png}&
\includegraphics[width=0.30\textwidth]{figures/percent_grid.png}&
\includegraphics[width=0.30\textwidth]{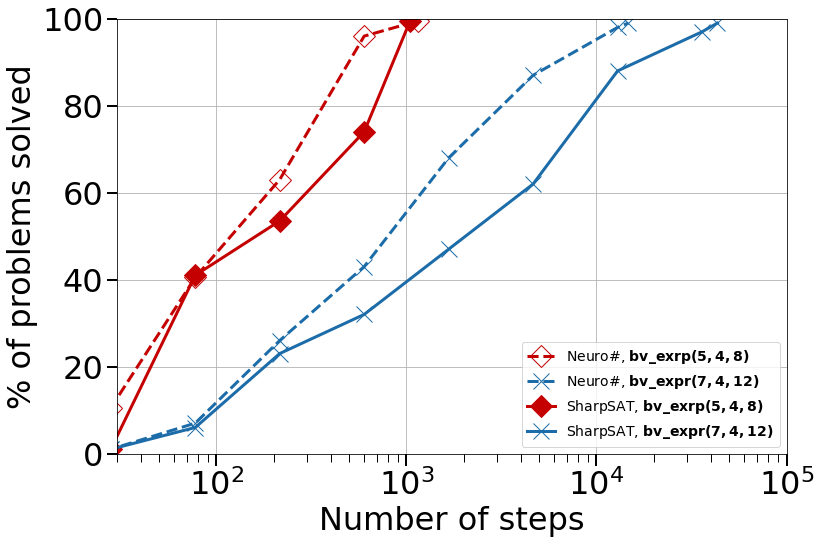}\\
(g) \cell[49] & (h) \grid{} & (i) \arith\\[0.4cm]

\multicolumn{3}{c}{
\begin{tabular}{cc}
\includegraphics[width=0.30\textwidth]{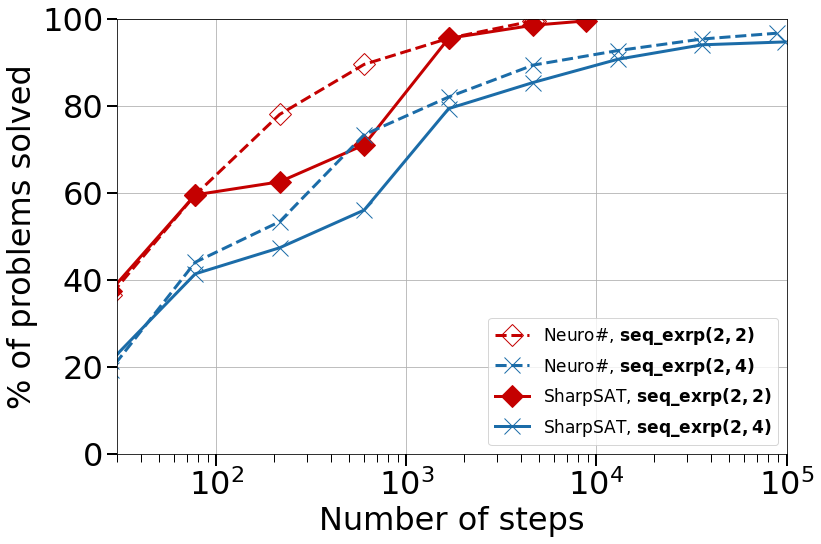}\\
(j) \itarith[]\\
\end{tabular}}

\end{tabular}

\caption{
\neurosharp generalizes well to larger problems on almost all datasets (higher and to the left is better). Compare the robustness of \neurosharp vs. \sharpsat{} as the problem sizes increase. Solid and dashed lines correspond to \sharpsat{} and \neurosharp, respectively. All episodes are capped at 100k steps.}
\label{fig:percent}
\end{figure*}

Figure \ref{fig:percent} compares the percentage of the problems solvable by \sharpsat{} vs. \neurosharp under a given number of steps. Notice the robustness of the learned model in \cell[35\&49] and \grid{}. As these datasets get more difficult, \sharpsat{} either takes more steps or completely fails to solve the problems altogether, whereas \neurosharp relatively sustains its performance. 

\subsection{Comparison with \ganak and Centrality}
\begin{figure}[tb]
\centering
\begin{tabular}{cc}
\includegraphics[width=0.45\columnwidth]{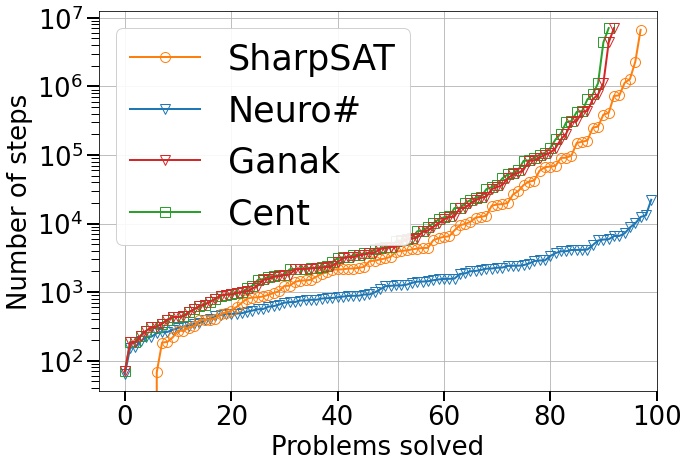}&
\includegraphics[width=0.45\columnwidth]{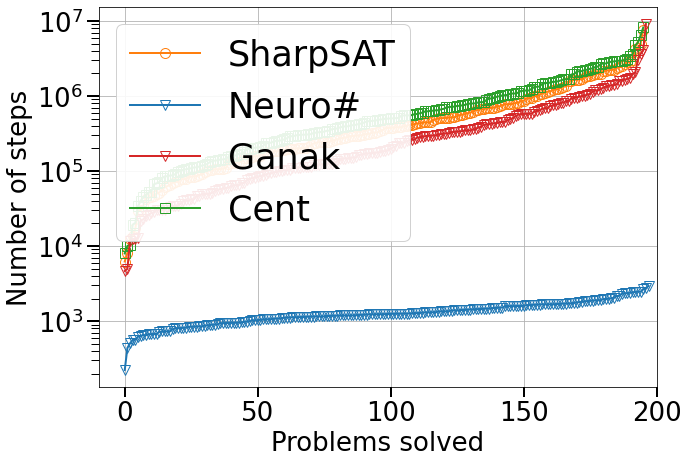}\\
\includegraphics[width=0.45\columnwidth]{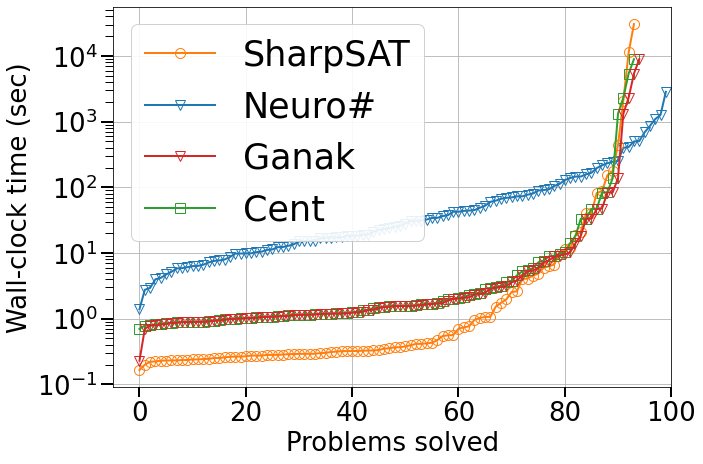}&
\includegraphics[width=0.45\columnwidth]{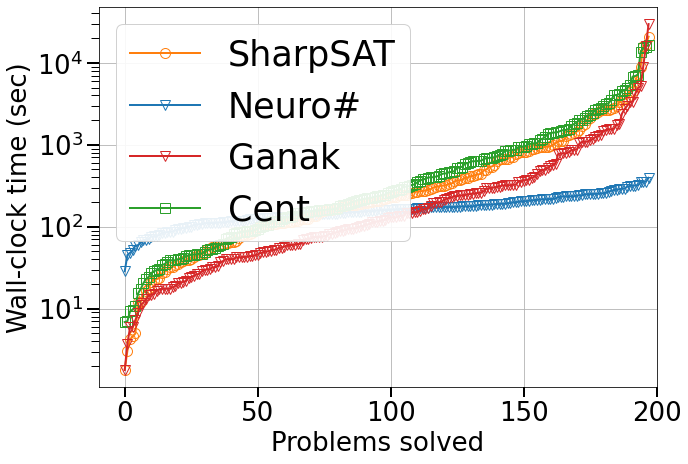}\\
(a) \cell[49,256,200] & (b) \grid[10,12]
\end{tabular}
\caption{Orders of magnitude reduction in the number of branching steps which translates to wall-clock improvements as problems get harder. Note, as explain in the paper, Python startup overhead skews results on easy problems.}
\label{fig:rebutt}
\end{figure}

\begin{figure*}[h]
\centering
\begin{tabular}{cc}
\includegraphics[width=7cm]{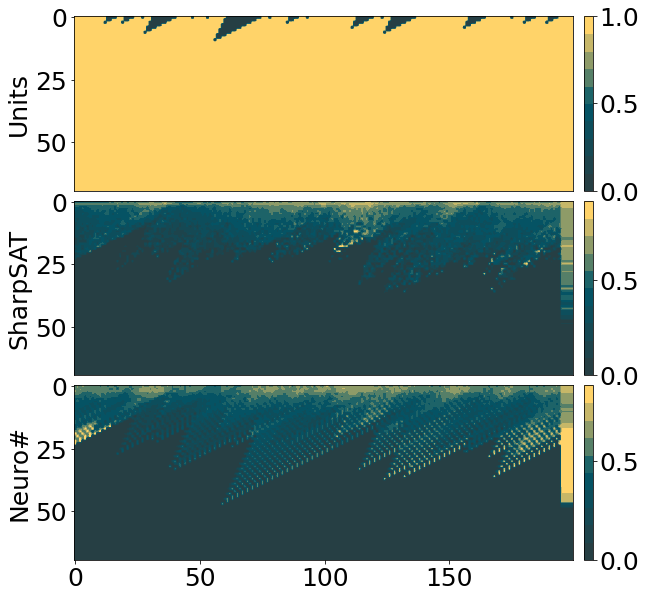}&
\includegraphics[width=7cm]{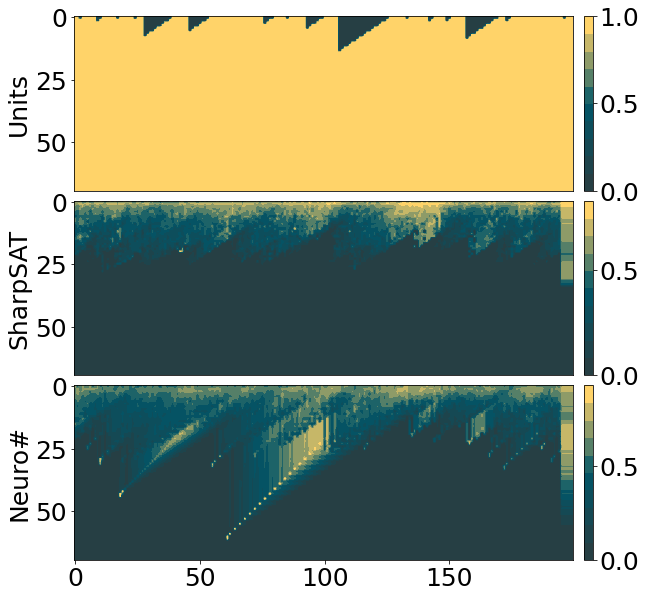}\\
(a) \cell[35,192,128]&(b) \cell[49,192,128]\\
\end{tabular}

\caption{Clear depiction of \neurosharp's pattern of variable branching. The ``Units'' plots show the initial formula simplification the solvers. Yellow indicates the regions of the formula that this process prunes. Heatmaps show the variable selection ordering by \sharpsat{} and \neurosharp. Lighter colours show that the corresponding variable is selected earlier on average across the dataset.}
\label{fig:heat_large_compare}
\end{figure*}

Figure \ref{fig:rebutt} shows the results of running \citeauthor{DBLP:conf/ictai/BliemJ19}'s centrality-based solver (henceforth ``Cent'')  and \ganak on the two datasets that we tested wall-clock time on (i.e., \cell{} and \grid{}). We observe that \sharpsat{}, Cent and \ganak behave more or less in the same performance regime, whereas \neurosharp deviates from the pack and emits the superlinear performance on the step counts. This results in wall-clock improvements for \neurosharp, which again happens as the problems get more and more difficult.

\begin{figure}[tbh]
\centering
\begin{tabular}{c}
\includegraphics[width=6cm]{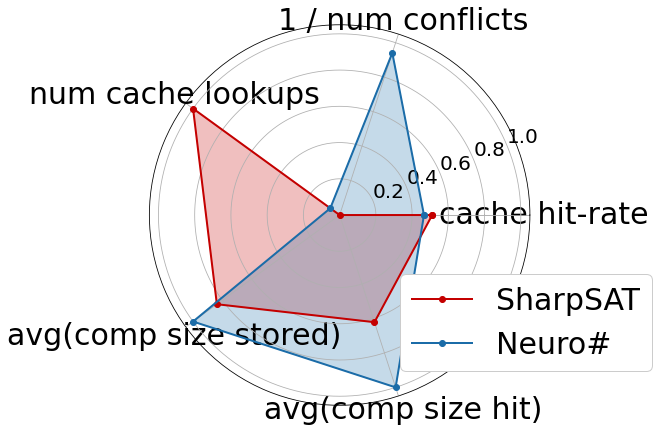}\\
(a) \cell[49,256,200]\\
\\
\includegraphics[width=6cm]{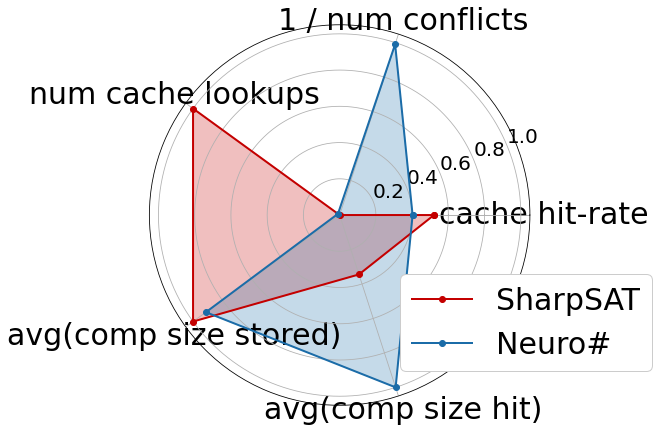}\\
(b) \grid[10,12]\\
\end{tabular}

\caption{Radar charts showing the impact of each policy across different solver-specific performance measures.}
\label{fig:radars_cell}
\end{figure}

\subsection{Discussion} 
In Section \ref{sec:discussion} we mentioned that, one of the main contributing factors to the upward generalization success of the model is its ability to observe similar components many times during training and we connected that to how effectively a problem gets shattered by the initial simplification (unit propagation) into smaller components. We visualized this phenomena for \cell{} via heat maps. In Figure \ref{fig:heat_large_compare} we provide full heat maps for larger problems of both \cell[35] and \cell[49]. Not only the ``shattering'' effect is evident from these plots, we can also observe that in both datasets \neurosharp branches on variables from the bottom going up. This matches with our conjecture presented in Section \ref{sec:discussion}.

\section{Trained Policy's Impact on Solver Performance Measures}\label{appendix:solver_analysis}
In this section we analyze the impact of \neurosharp on solver's performance through the lens of a set of solver-specific performance measures. These measures include: 
\begin{enumerate*}[label=\textbf{\arabic*.}]
\item Number of conflict clauses that the solver encounters while solving a problem (\textproc{num conflicts}), \item Total (hit+miss) number of cache lookups (\textproc{num cache lookups}), \item Average size of components stored on the cache (\textproc{avg(comp size stored)}), \item Cache hit-rate (\textproc{cache hit-rate}) and \item Average size of the components that are successfully found on the cache (\textproc{avg(comp size hit)}) \end{enumerate*}.

A conflict clause is generated whenever the solver encounters an empty clause, indicating that the current sub-formula has zero models. Thus the number of conflict clauses generated is a measure of the amount of work the solver spent traversing the \emph{non-solution space} of the formula. Cache hits and the size of the cached components, on the other hand, give an indication of how effectively the solver is able to traverse the formula's \emph{solution space}. In particular, when a component with $k$ variables is found in the cache (a cache hit) the solver does not need to do any further work to count the number of solutions over those $k$ variables. This could potentially save the solver $2^{O(k)}$ computations.
This $2^{O(k)}$ worst case time is rarely occurs in practice; nevertheless, the number of cache hits, and the average size of the components in those cache hits give an indication of how effective the solver is in traversing the formula's solution space. 
Additional indicators of solver's performance in traversing the solution space are the number of components generated and their average size. Every time the solver is able to break its current sub-formula into components it is able to reduce the worst case complexity of solving that sub-formula. For example, when a sub-formula of $m$ variables is broken up into two components of $k_1$ and $k_2$ variables, the worst case complexity drops from $2^{O(m)}$ to $2^{O(k_1)} + 2^{O(k_2)}$. Again the worst case rarely occurs (as indicated by the fact that \#SAT solvers do not display worst case performance on most inputs), so the number of components generated and their average size provide only an indication of the solver's effectiveness in traversing the formula's solution space.

In Figure \ref{fig:radars_cell} we plot these measures for \cell[49,256,200] and \grid[10,12]. Looking at the individual performance measures, we see that the \neurosharp encounters fewer conflicts (larger \textproc{1/num~conflicts}), meaning that it is traversing the non-solution space more effectively in both datasets. The cache measures, indicate that the standard heuristic is able to traverse the solution space a bit more effectively, finding more components (\textproc{num cached lookups}) of similar or larger average size. However, \neurosharp is able to utilize the cache as efficiently (with comparable cache hit rate) while finding components in the cache that are considerably larger than those found by the standard heuristic. In sum, the learnt heuristic finds an effective trade-off of learning more powerful clauses, with which the solver can more efficiently traverse the non-solution space, at the cost of a slight degradation in its efficiency traversing the solution space. The net result in an improvement in the solver's run time.

\section{More Ablations}\label{appendix:ablation}
\paragraph{Variable Score} We mentioned in Section \ref{sec:background} that \sharpsat{}'s default way of selecting variables is based on the VSADS score which incorporates the number of times a variable $v$ appears in the current sub-formula, and (a function of the) number of conflicts it took part in. At every branching juncture, the solver picks a  variable among the ones in the current component with maximum score and branches on one of its literals (see Algorithm \ref{alg:sharp_dpll}). As part of our efforts to improve the performance of our model, we performed an additional ablation study over that of Section \ref{sec:results}. Concretely, we measured the effect of including the variable scores in our model. We start with a feature vector of size $2$ for each literal, and pass it through an MLP of dimensions $2\times32\times32$ to get the initial literal embedding.
We tested on \cell[49,256,200] and \grid[10,12] datasets (Figures \ref{fig:cell_ablation_appendix} \& \ref{fig:ablation_appendix}). For both datasets, the inclusion of the variable scores produced results inferior to the ones achieved without them! This is surprising, though consistent with what was observed in \cite{LedermanRSL20}.

\begin{table}[h]
  \centering
    \includegraphics[width=0.40\textwidth]{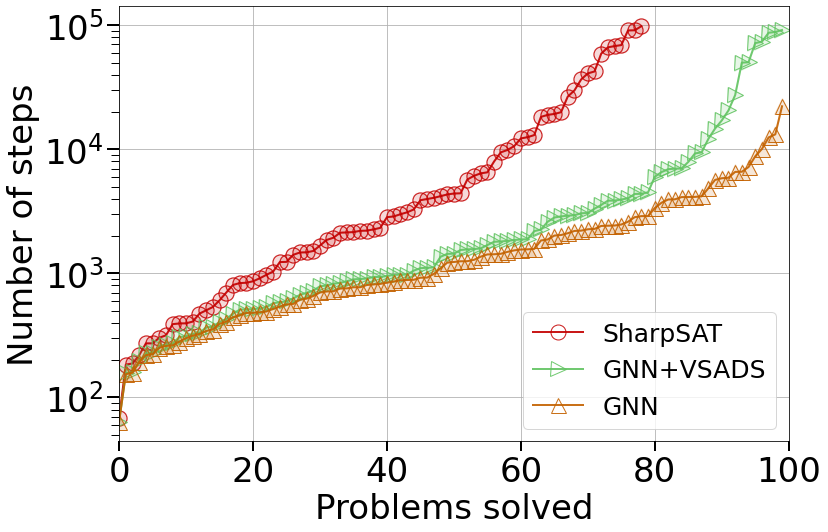}
    \captionof{figure}{Cactus Plot -- Inclusion of VSADS score as a feature hurts the upward generalization on \cell[49,256,200] (lower and to the right is better). A termination cap of 100k steps was imposed on the solver.}
    \label{fig:cell_ablation_appendix}
    
\vspace{1em}

    \includegraphics[width=0.40\textwidth]{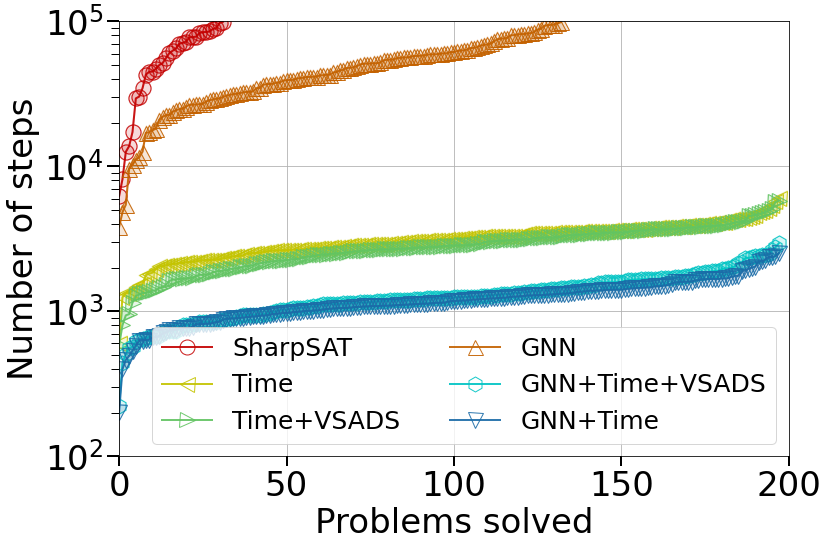}
    \captionof{figure}{Cactus Plot -- Ablation study on the impact of the ``time'' and VSADS features over upward generalization on \grid[10,12] (lower and to the right is better). A termination cap of 100k steps was imposed on the solver.}
    \label{fig:ablation_appendix}
\end{table}

\end{document}